\documentclass[leqno]{article}

\usepackage[preprint]{neurips_2023}

\usepackage[utf8]{inputenc} 
\usepackage[T1]{fontenc}    
\usepackage{hyperref}      
\usepackage{url}            
\usepackage{booktabs}      
\usepackage{amsfonts}     
\usepackage{nicefrac}       
\usepackage{microtype}  
\usepackage{xcolor}    
\usepackage{graphicx}
\usepackage{subcaption}
\usepackage[fleqn]{amsmath}
\usepackage{enumitem}

\title{Pure Planning to Pure Policies and In Between with a Recursive Tree Planner}

\author{%
  A. Norman Redlich\\
  ANR Robot \\
  \texttt{norman@anrrobot.com} \\
}

\begin{document}

\maketitle

\begin{abstract}
A recursive tree planner (RTP) is designed to function as a pure planner without policies at one extreme and run a pure greedy policy at the other. In between, the RTP exploits policies to improve planning performance and improve zero-shot transfer from one class of planning problem to another. Policies are learned through imitation of the planner. These are then used by the planner to improve policies in a virtuous cycle. To improve planning performance and zero-shot transfer, the RTP incorporates previously learned tasks as generalized actions (GA) at any level of its hierarchy, and can refine those GA by adding primitive actions at any levels too. For search, the RTP uses a generalized Dijkstra \citep{dijkstra} algorithm which tries the greedy policy first and then searches over near-greedy paths and then farther away as necessary. The RPT can return multiple sub-goals from lower levels as well as boundary states near obstacles, and can exploit policies with background and object-number invariance. Policies at all levels of the hierarchy can be learned simultaneously or in any order or come from outside the framework. The RTP is tested here on a variety of Box2d \citep{cato} problems, including the classic lunar lander \citep{farama}, and on the MuJoCo \citep{mujoco} inverted pendulum.
\end{abstract}

\section{Introduction}

There are various ways that a robot can exploit knowledge of its world to make decisions about which actions it should take to achieve its goals. One approach is to equip the robot with an inference engine together with knowledge of the world in the form of a dynamics model. The dynamics model, or transition function, predicts what state $s’$ the world will be in if the robot takes action $a$ in the current state $s$: $s’=f(a,s)$. The inference engine uses this model to search through possible futures to find a series of actions, a plan, that will lead the robot to its goal. At the opposite extreme, the robot does not know a dynamics model and does not have an inference engine. Instead, the robot uses reinforcement learning (RL) while interacting with the robot’s environment to learn what action $a$ in state $s$ is most likely to get the robot closer to achieving its goal. Knowledge in this case is in the form of a policy, $p(a|s)$, which can be a stochastic function whose maximum over a is the best action to take.

This is of course an artificial dichotomy, since it can be helpful to use both types of knowledge. In this paper, I explore one way to exploit both types together in the same system. The robot will start with a pure planner, consisting of an inference engine and a dynamics model or transition function encompassing a large set of problems. The pure planner should be able to solve any of these problems, but in practice this may not be practical with realistic computing resources and time. Instead, the pure planner may be used to solve a few simpler examples of some problem, or will be used to solve related but easier problems or sub-problems. The solutions, in the form of plans, will then be used as data for a supervised learning algorithm —imitating the planner — which learns stochastic policies. The planner is devised to be able to exploit these policies to narrow its search over possible plans, and hence speed up and improve the accuracy of its planning. This policy-informed planner then generates more problem solutions, and so on, thus iteratively improving the policies (similar to \citep{anthony}).

After sufficient iterations of this plan-learn-plan (PLP) framework, the policies may be able to stand on their own. In this case, instead of searching, the planner follows the greedy policies, those which maximize $p(a|s)$. Hence, in this framework we need a planner that can work as a pure planner at one extreme, then exploit policies as they are learned to improve planning, and finally run those policies in pure greedy mode at the opposite extreme.

Although not explored fully here, the planner should also be able to use policies that were learned directly by interacting with the environment using RL or any other approach, not from imitating the planner. This closes the loop.

The primary purpose of this paper is to design one particular such planner, and to use it to explore the spectrum from pure dynamics to pure policies. The goal is to have our cake and eat it too. Planners are typically slow but good at finding solutions to novel problems and they require no outside experience. Polices are fast, but typically don’t generalized well and often require lots of problem examples (here) or outside experience.

The planner introduced here is a forward search tree planner which can call itself recursively and hence is hierarchical. Forward search allows goal states to be discovered by the planner since they are often not known in advance. Hierarchy allows the search space to be sampled sparsely at higher levels by using lower level planners to take large steps in state space. The higher level planners can use both primitive actions which call the transition function directly and/or \textit{generalized actions} which recursively call lower level planners. Each generalized action has its own policy and sub-goals which it provides to a sub-planner to solve a sub-problem. For example, sub-problems might be simple $grab$ and simple $place$ tasks which a higher level planner uses to solve a pick and place problem.

The tree planner uses a generalized generalized Dijkstra algorithm \citep{dijkstra,lavalle} which can try \textit{any} next action from \textit{any} state in the search tree in any order. When a policy is available, this allows the search tree to try the depth-first path corresponding to the greedy policy first. The planner can then use the stochastic policy to try the next most probable action and state. In this way, the planner searches in the neighborhood of the greedy policy, which I will call \textit{near-greedy} search, and then further and further away. When no policy is available, the planner can search in any order desired or randomly pick actions and states, which is often a good choice.

This recursive tree planner (RTP) turns out to be good at generalizing from one class $C_i$ to another $C_j$, which I will refer to as zero-shot transfer. Both hierarchy and near-greedy search contribute to this. The planner can also exploit boundary states (as in RRT,  \citep{lavalle} ), and can return multiple sub-goals from a single search which speeds hierarchical planning. It is also capable of simultaneous multi-level policy learning, and can take advantage of background and object-number invariance.

The RTP algorithm is domain independent in the sense that the search algorithm is not modified for different problems. Domain dependence is kept wrapped inside black-box calls to specific transition and reward functions, and to stochastic policies. The domain dependent implementation details, such as which action and state representations to use for policies, have a significant impact on performance, and are discussed in this paper. The RTP algorithm for discrete actions and states is given first, but is then extended to continuous actions and states. The transition function — dynamics — in this paper is deterministic for all problems.

\section{Related work}
\label{related}

The PLP framework here expands on the approach in \citep{anthony} where they iteratively run a planner with a policy, and then learn to imitate the resulting plans to improve the policy. They apply their approach to improve performance of an MCTS planner. This is very similar to alpha-zero \citep{silver} where policies are improved by an MCTS planner and then supervised learning is used to update the policy. As in \citep{anthony} , this exploits the fact that planners are good at exploration while supervised learning is robust at function approximation. Here, the framework in \citep{anthony}  is extended to plan and learn for one class of problems, and then apply the policies to other classes: zero-shot transfer. 

Moreover, rather than apply the framework to an existing planner, I design the planner specifically to explore how planning and policies interact, and in particular how combining them can lead to zero-shot transfer. This is similar in spirit to the Dyna program of \citep{sutton}, which explores how to learn both dynamics and policies from the environment and also how best to balance dynamics and policies in problem solving. For now, I only explore the latter question: given known dynamics, what are the best ways to incorporate policies and dynamics into the same system?

One result, here, is that hierarchy is extremely beneficial for both zero-shot transfer and for improving in-class planning too, especially for long horizons. In one respect, this bolsters the conclusion in (\citep{nachum} that hierarchies improve exploration, since that is clearly seen here. Of course, hierarchical planning has a long history, see \citep{bercher} for a survey. Often these planners use pre-existing libraries, or mixed techniques, such as combined task and motion planning  \citep{wang,dearden}. Refinement often increases as one moves down the hierarchy as discussed in (\citep{ghallab}, chapter 3). On the other hand, the planner here uses the same architecture at all levels, and refinement can be added at any level by using primitive actions to increase resolution and explore new regions of configuration space. The approach used here, for good or bad, doesn’t distinguish between task and motion planning, using the same planner for all problems.

The benefit of returning multiple sub-goals in a tree planner is seen in the hierarchical MCTS planner in \citep{vien}, and both that paper and \citep{bai} use state abstraction much like the generalized actions here. MCTS planners can handle non-deterministic dynamics, unlike the planner here, but exploring hierarchical planning is easier without uncertainty, so I stick to deterministic dynamics for now.

Another property of the RTP planner, here, which contributes to zero-shot transfer is the return of boundary states which helps the planner explore configuration space efficiently. This is similar to the use of boundary states by algorithms such as RRT \citep{lavalle}, although the state space sampling here is done by low level planners in place of RRT sampling.

The planner here naturally allows multi-level simultaneous policy learning which can be used to solve problems similar to those using hierarchical reinforcement learning in \citep{levy}. Actually, policies at all levels of the RTP hierarchy can be learned simultaneously or asynchronously, and they can be learned from a planner solving one problem class, then fine-tuned for a different problem class. This is because all that is needed to learn a policy is action-state data from within the RTP hierarchy.

Another approach with a long history is hierarchical reinforcement learning \citep{sutton,dietterich,parr} where options are the equivalent of the generalized actions — sub-planners — here. If dynamics are not known, then some form of reinforcement learning may be the only choice. However, if a model is known, then I believe it should to be exploited. One reason for this is that planners solve particular problem examples one at a time and a particular problem can be much simpler to solve than a class of problems. The action-state data from these simpler planning problems is aggregated here to learn a global policy for the class. This is in contrast to RL where the global policy for the entire class is used to both solve individual problems and to learn simultaneously.

In place of policies, another way to incorporate additional knowledge into planners, beyond dynamics, is to use heuristics such as A*, see for example \citep{lavalle,ghallab}. There are a few differences between heuristics and the policy approach here. The first is that a policy learned through the PLP framework might be good enough to stand on its own without the planner. The second is that policies can come from some other source, such as RL. Third, they also do not require advanced knowledge about goal states. Fourth, policies can be very flexibly incorporated into a planner in the form of generalized actions at multiple levels and mixed in with primitive actions. This contributes to zero-shot capability.

Finally, forward search and generalized Dijkstra have a long history, described very nicely in the book \textit{Planning Algorithms} \citep{lavalle}. The particular variant used here, where the global policy is used to pick both an action and a state to try next is a little different than the generalized Dijkstra algorithm in (\citep{lavalle}, p. 47) where states can be chosen any way you like, but \textit{all} allowed actions from each state are tried before choosing the next state. I expect that the version here has been discussed previously, but I haven't found a reference.

\section{The Plan Learn Plan (PLP) Framework}
\label{prob_sys_form}

\subsection{Problem Formulation 1}
\label{prob_form_1}

This formulation is for a finite set of discrete actions and states which simplifies the planning algorithms in Section~\ref{the_planner}. A second formulation, \ref{prob_form_2}, will add back continuous actions and states while modifications to the algorithms needed for \ref{prob_form_2} are discussed in \ref{continuous}. In this paper, the dynamics are deterministic.

Let’s define a world W in which a very large set $C_W$ of problems classes $C$ share the same state space and dynamics:

\begin{equation} 
\label{W}
W=\{S,U,f,C_W\}, \quad C=\{r,S_G\}, \quad C \in C_W \\
\end{equation}

where 

\begin{itemize}[leftmargin=*]
\item $S$ is a finite discrete state space.
\item$U(s)$ is a finite set of allowed actions for each state $s \in S$. $U(s) \in U$ is a sub-set of all primitive actions $U$.
\item$f(u,s) \in S$ is a state transition function for every action $u$ and state $s$: $s’=f(u,s)$.
\item$C_W$ is the set of all problem classes that share $S$,$U$, and $f$.
\item$r(u,s)$ is a reward function for each class $C$ where $r(u,s)$ is the reward received in state $s’=f(u,s)$ after taking action u.
\item$S_G$ is a set of goals states for each class $C$.
\item$s_I \in S$ is an initial state for each problem example.
\end{itemize}

In addition, I define:

\begin{itemize}[leftmargin=*]
\item $\pi= \{(u_0,s_0),(u_1,s_1),…\}$ is a plan which is a linked set of action-state pairs where $s_{i+1}=f(u_i,s_i)$
\item $R( \pi )$ denotes the cumulative reward adding up $r(u_k,s_k)$ for all $(u_k,s_k)$ in $\pi$. $R(s)$ will denote the cumulative reward inside a search tree from the root up to state $s$.
\item $S_B$ are boundary states which are allowed states near to obstacles or other forbidden states.
\end{itemize}

The goal states $S_G$ do not need to be explicitly known in advance. They can be discovered by the system, for example by receiving a very large reward. However, I will leave this out of the formulation and consider it an implementation detail. Similarly, $U(s)$ may not be explicitly known in advance for each $s$, but may be implemented by not allowing the system to transition into an obstacle or other forbidden state.

The transition function can be a dynamics simulator such as box2d \citep{cato} or MuJoCo \citep{mujoco}. The simulator is required to be deterministic and be capable of reproducibly returning the next state from any system state.   

$S_B$ are are easily discovered by plans and sub-plans as the states encountered just before an attempted transition into an obstacle or other forbidden state. They often have high sampling value in searching for a plan. Each problem example, $s_I$, may introduce different obstacles and other forbidden states.

\subsection{Problem Formulation 2}
\label{prob_form_2}

Extending Formulation 1 to continuous actions and states, the following changes and additions are made:

\begin{itemize}[leftmargin=*]
\item $S$ is a continuous state space.
\item $U(s)$ is a continuous action space.
\item $e(s,s’)$ is an equals function for each problem class $C$. $e(s,s’)=true$ when $s$ and $s’$ are within a minimal sampling distance from each other. This is a domain dependent function like $r(u,s)$ but is a black box to the planning algorithm.
\end{itemize}

\begin{figure}
  \includegraphics[width=0.9\linewidth]{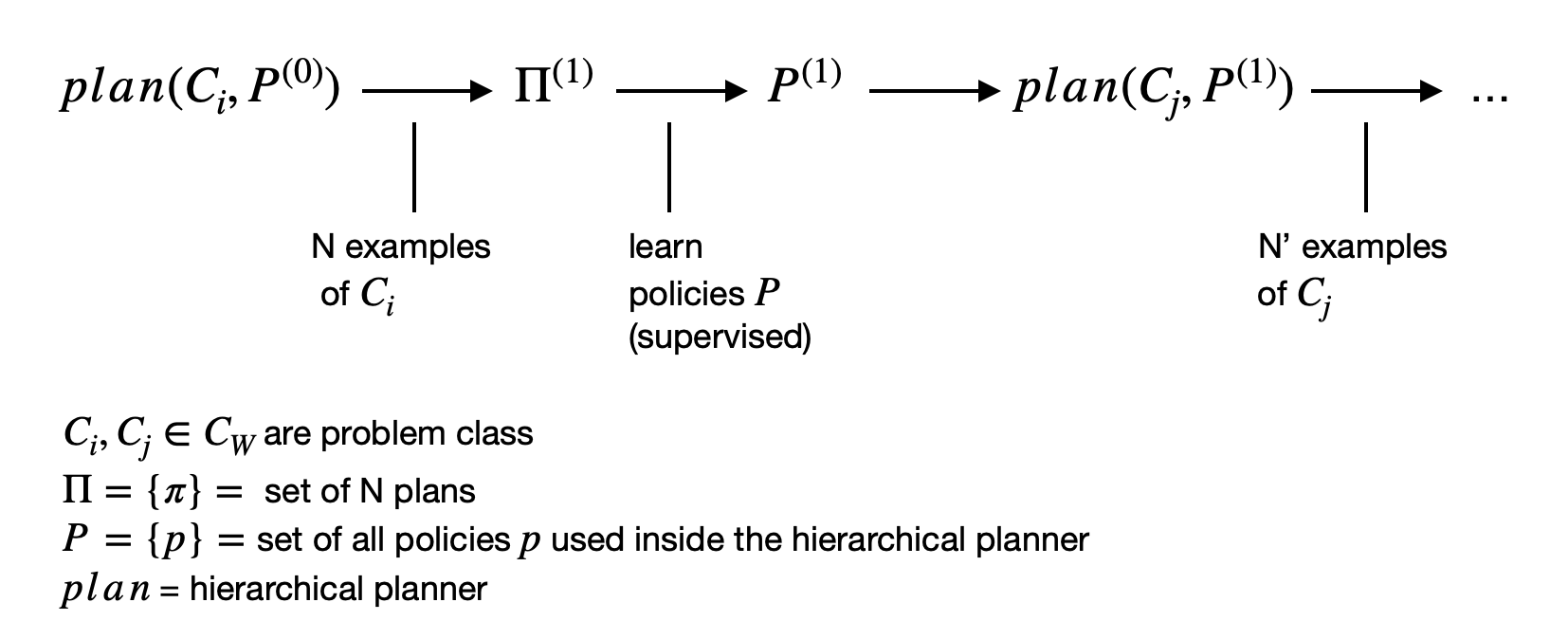}
  \caption{Plan Learn Plan (PLP) framework: planner data from class $C_i$ is used to learn policies $P^{(0)}  \to P^{(1)}$. Bootstrapping continues to improve the policies for class $C_i$ when $C_j=C_i$, or policies can be used for zero-shot solutions to $C_j$. Learning can continue to improve the policies for $C_j$ too. $P^{(0)}$ can be non-learned initial search strategies for pure planning.}
  \label{framework}
\end{figure}

\subsection{Plan Learn Plan (PLP) Framework}
\label{PLP}

The general framework is shown in Figure~\ref{framework}. First, a hierarchical planner, $plan(C_i,P)$, is called N times to find a set of plans, $\Pi$, for the set of learning examples from class $C_i$. The hierarchical planner uses a set of policies $P=\{p\}$, one $p$ for each sub-planner, and one $p$ for the highest level. Note that there can be more than one policy at each level. The initial policies $P^{(0)}$ are not learned: they can be uniform or random, for example. The batch of planning solutions,  $\Pi$, is then fed into \textit{supervised learning} algorithms to learn an updated set of policies $P^{(1)}$. The planner can then run again on a set of examples from $C_i$, but this time speed and/or accuracy should have improved. The new or total set of plans can then be used to learn better policies and so on. When $C_j=C_i$ this is pretty much the same as the iterative framework in \citep{anthony}.

Optionally, the policies can be used by the planner for zero-shot transfer to solve problems of a different class $C_j$. Further bootstrap learning can continue for $C_j$ too. Not shown is that policies can be spun out of the framework when they are good enough to be used without the planner: for example, to initialize RL. Alternatively, policies learned from the outside, for example from RL, can be incorporated back into the framework.

\section{The Planner}
\label{the_planner}

\subsection{Design Goals and Properties}
\label{design_goals}

At a high level the planner needs to:

\begin{itemize}[leftmargin=*]
\item Use policies to improve planning.
\item Run as a pure planner at one extreme and run the pure greedy policy at the other. 
\item Perform zero-shot transfer using policies from class $C_i$ to solve $C_j$.
\end{itemize}

To achieve these goals, the planner will have the following properties. These will be described in detail below, and their contribution to achieving the goals will be tested by the experiments in Section~\ref{experiments}:

\begin{enumerate}[leftmargin=*]
\item Hierarchy: it can recursively call itself.
\item Near-greedy search: given a stochastic policy, the planner should try the greedy policy first and then search near this policy, and then farther and farther away.
\item Generalized actions: re-use of policies inside the hierarchy as generalized actions.
\item Boundary states: return boundary states near objects or other forbidden states.
\item Multiple sub-goals: sub-planners can return multiple sub-goals within the hierarchy.
\item Refinement at all levels: primitive actions can be combined with generalized.
\item Simultaneous or asynchronous policy learning within the hierarchy.
\item Invariance: background and object number invariance in policies can be exploited by the planner.
\end{enumerate}

\begin{figure}
  \centering
  \includegraphics[width=\linewidth]{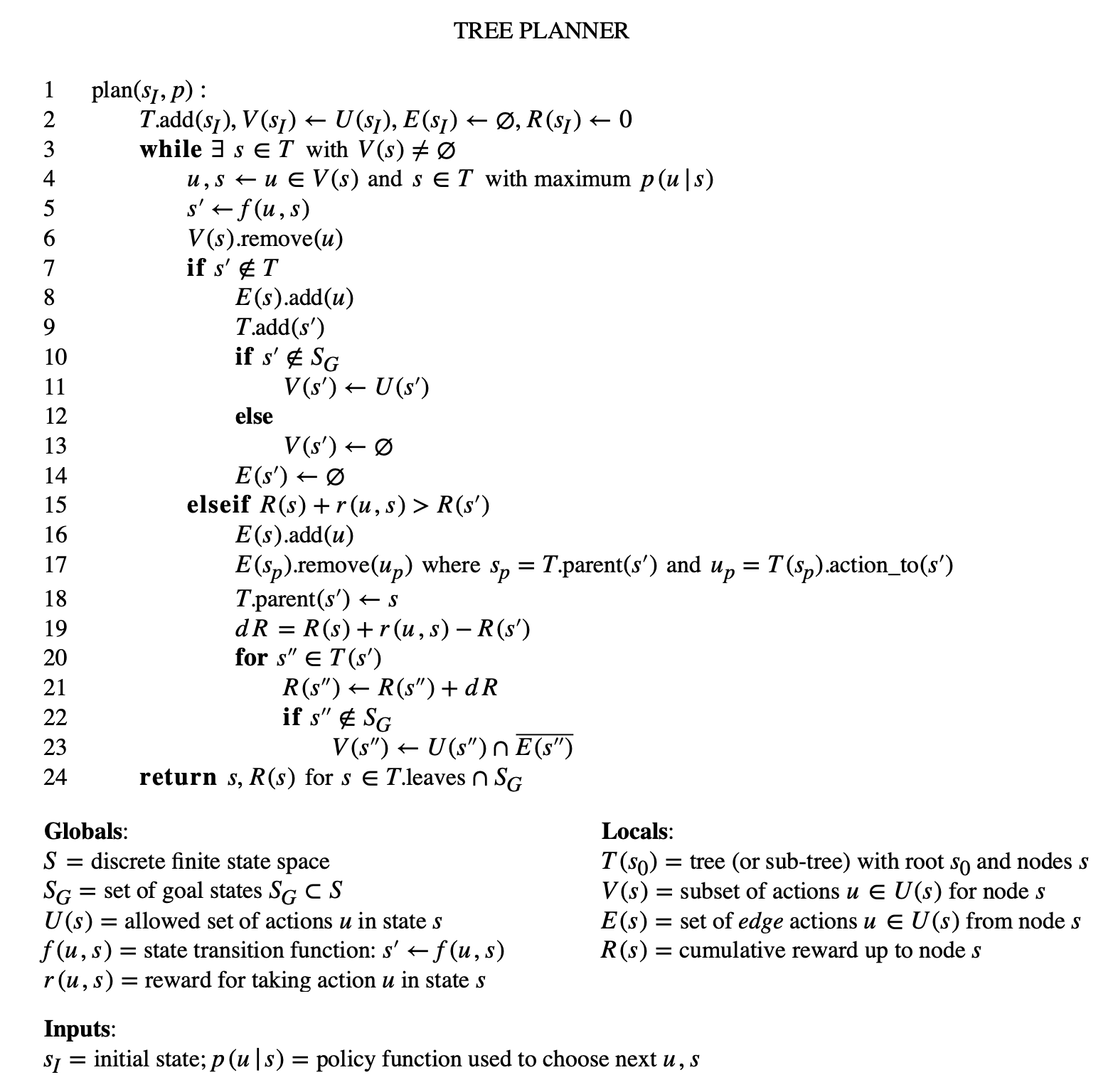}
  \caption{Pseudocode for the non-hierarchical planner. Line 4 implements near-greedy search. If a next state $s'$ is already in $T$ and adding it to $s$ improves its $R(s')$, line 15, then it is moved along with its subtree to $s$, lines 16-23.}
 \label{fig_TP}
\end{figure}

\subsection{The Tree Planner (TP)}
\label{TP}

Let’s start with the non-hierarchical planner, TP in Figure~\ref{fig_TP}, and later use this planner in Section~\ref{RTP} to build a hierarchical version. The planner is designed for discrete actions and states, formulation \ref{prob_form_1}, but later, in \ref{continuous}, generalized to continuous actions and state. 

The planner is designed to find an \textit{optimal} -- max $R(s_G)$ -- path from start state $s_I$ to goal state $s_G$, given a full search over all actions and states that can be encountered from the start state $s_I$. If there isn’t time for a full search, then the planner should at least monotonically improve $R(s_G)$. When path rewards are zero -- R(s)=0 -- the planner should find a \textit{feasible} path to $s_G$.

To these ends, I chose a forward search tree planner using a generalized Dijkstra algorithm as the starting point \citep{lavalle}. Forward search was chosen because the goal states may not be known in advance. A generalized Dijkstra algorithm, such as described in  \citep{lavalle}, keeps the search tree small by allowing only one node per states, so search speed is proportional to the number of states. Otherwise, the tree could grow exponentially.

However, there is one significant change which needs to be made to the algorithm in \citep{lavalle}: That algorithm allows states in the tree to be expanded in any order, but each state is \textit{fully} expanded by trying \textit{all} allowed actions from that state. On the other hand, here, we need to be able to try the greedy policy first. Hence, the algorithm needs to be modified to be able to try both states and actions in any order.

I believe this is achieved in the algorithm shown in Figure~\ref{fig_TP}. The tree $T(s_I)$ with root $s_I$ consists of nodes which are states $s \in S$. The algorithm keeps track of the set, $E(s)$, of actions $u$ which correspond to existing edges in $T$ from state $s$ to child state $s'$ obtained using $s’=f(u,s)$. The algorithm also keeps track of the set $V(s)$ of actions $u$ that are still available to be tried from node $s$.

As promised, the tree is expanded, line 4, using the stochastic policy $p(u|s)$. The next state $s$ and action $u$ to try are the ones that maximize the policy $p(u|s)$.

\begin{equation} 
\label{max_p}
u,s = argmax_{u,s} \; p(u|s) \quad s.t. \; u \in V(s)
\end{equation}

over all nodes $s$ in the existing tree $T$ and all available actions $u \in V(s)$ from $s$. It may seem strange to maximize over the conditional variable $s$, but Equation~\ref{max_p} will try the greedy path $\pi*=\{u*,s*\}$  first as long as $p(u*|s*)$ is strongly peaked along the greedy path. The actual requirement is

\begin{equation} 
\label{pp_greedy}
p(u*|s*) > p(u’|s’),  \quad \forall \;greedy \; (u*,s*) \; and \; \forall \; nongreedy \; (u’,s’) \; \in T
\end{equation}

In practice, I find that this is frequently satisfied. Once the greedy path has been tried, Equation~\ref{max_p} chooses the next closest to greedy state and available action, and so on.

To find optimal paths, the algorithm moves sub-trees around whenever a better path is found to an existing state $s’$ in the tree, see line 15. This is the case when the new path through $s$ to $s’$ has cumulative reward, $R(s)+r(u,s)$, which is greater than $R(s’)$ for the current path to $s’$. Then $s’$ is detached from its current parent and attached to $s$ instead. In addition, any sub-tree flowing downstream from $s’$ is moved along with $s’$. The rest is bookkeeping: cumulative rewards in the sub-tree need to be updated, and, most importantly, actions from states in the sub-tree which failed line 15 in the past need to be made available in $V(s)$ to try again. This last step is required because these sub-tree states now have larger cumulative rewards so the next time line 15 is tried they may win.

One side benefit of the algorithm is that it optimizes the paths to all states in the tree, not just the goal states $s_G$.

The algorithm returns the set of leaf states s which are also goal states, $S_G$, along with their cumulative rewards $R(s)$. Because each state in the tree has a unique parent, a unique plan $\pi$ for each goal state can be found by backtracking to the root, although this is not explicitly shown in the figure.

\subsection{Optimality}
\label{optimality}

An informal proof of the algorithm’s optimality is given in Appendix B (I assume that a proof is known, but I haven't found a reference). A sufficient condition for optimality is shown to be negative rewards for all transitions except for the transition to the goal state which can be any value. This negative reward condition is equivalent to the Dijkstra algorithm requirement of positive costs (costs being the negative of rewards).

When rewards are not all negative, the algorithm needs to be modified in line 15 of Figure~\ref{fig_TP} to avoid potential loops. With this modification, positive rewards can be used and the algorithm will improve every time line 15 is true, but there is no guarantee that a globally optimal solutions will be found.

Often it is desirable to place limits on the maximum size of the search tree or on the maximum search depth. In this case, the algorithm is again not guaranteed to find an optimal plan, but it will improve monotonically because there is no way that a state can be removed from the tree, and line 15 is the only mechanism for changing the cumulative rewards of states: rewards for the state $s’$ and the downstream states can only go up. When the search is guided by a policy, an optimal or near-optimal plan is still often found even without a complete search, although this is not guaranteed.

A near-greedy policy-guided search works even better for finding feasible plans since one can stop searching as soon as one is found.

\begin{figure}
 \centering
  \includegraphics[width=0.9\linewidth]{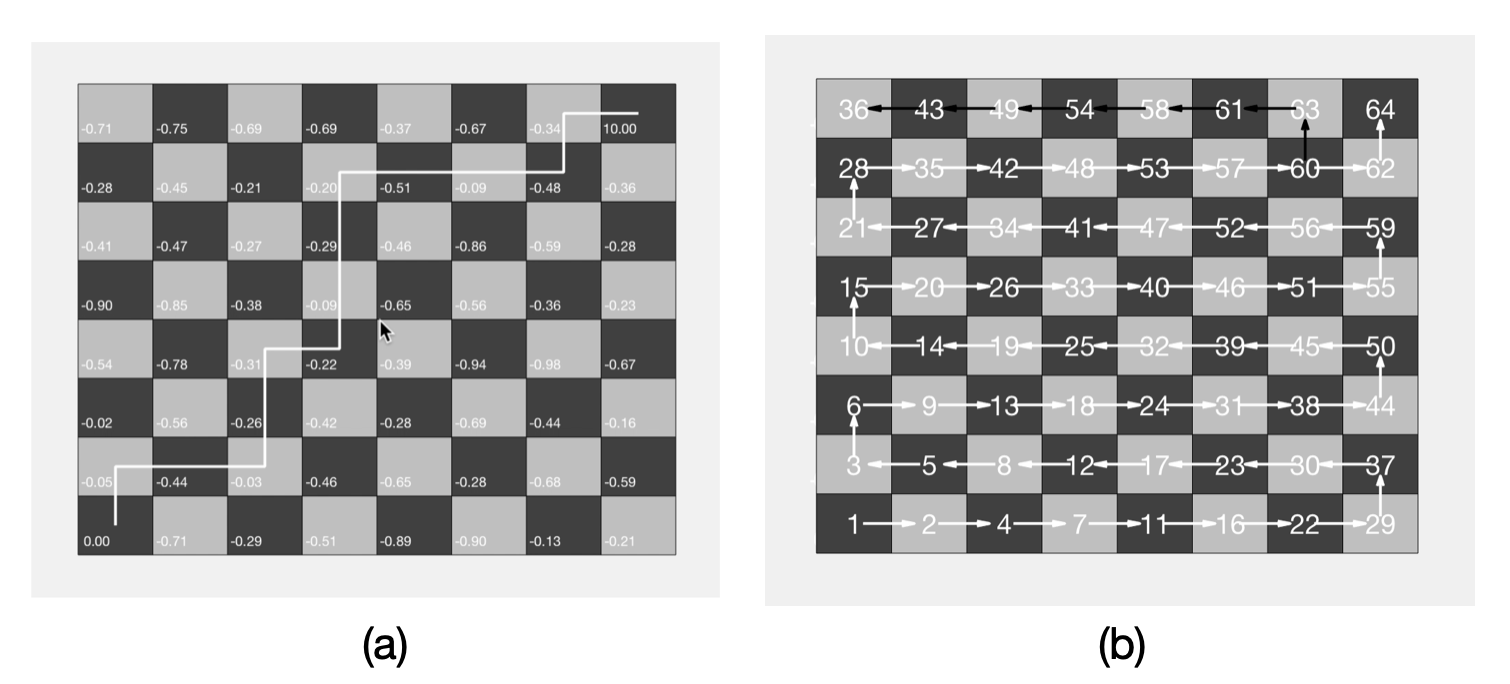} 
  \caption{Examples of optimal planning solutions using the TP, Section~\ref{TP}. (a) Optimal path for random negative rewards, shown and random choice of next action and state. (b) Optimal path for uniform positive rewards using the TP with breadth first.}
\label{checkerboards}
\end{figure}

I tested the algorithm on a checkerboard with a random set of negative rewards, as shown in Figure~\ref{checkerboards}a. The solution is the same as the one found using an exhaustive search, however, the exhaustive search uses a search tree of size $2*10^6$ and takes 30 seconds on my computer, while the TP algorithm uses a tree of size 64 (the number of states) and takes 0.03 seconds. Most importantly, this same solution is found for any random or systematic order used for states and actions.

I also tested the algorithm, adding the no-loop constraint, with uniform positive rewards and it correctly produces the maximal length plan shown in Figure~\ref{checkerboards}b. There is, however, no guarantee of optimality in this case, so when the search is random, a good but sub-optimal solution was found, not shown.

\begin{figure}
 \centering
   \includegraphics[width=\linewidth]{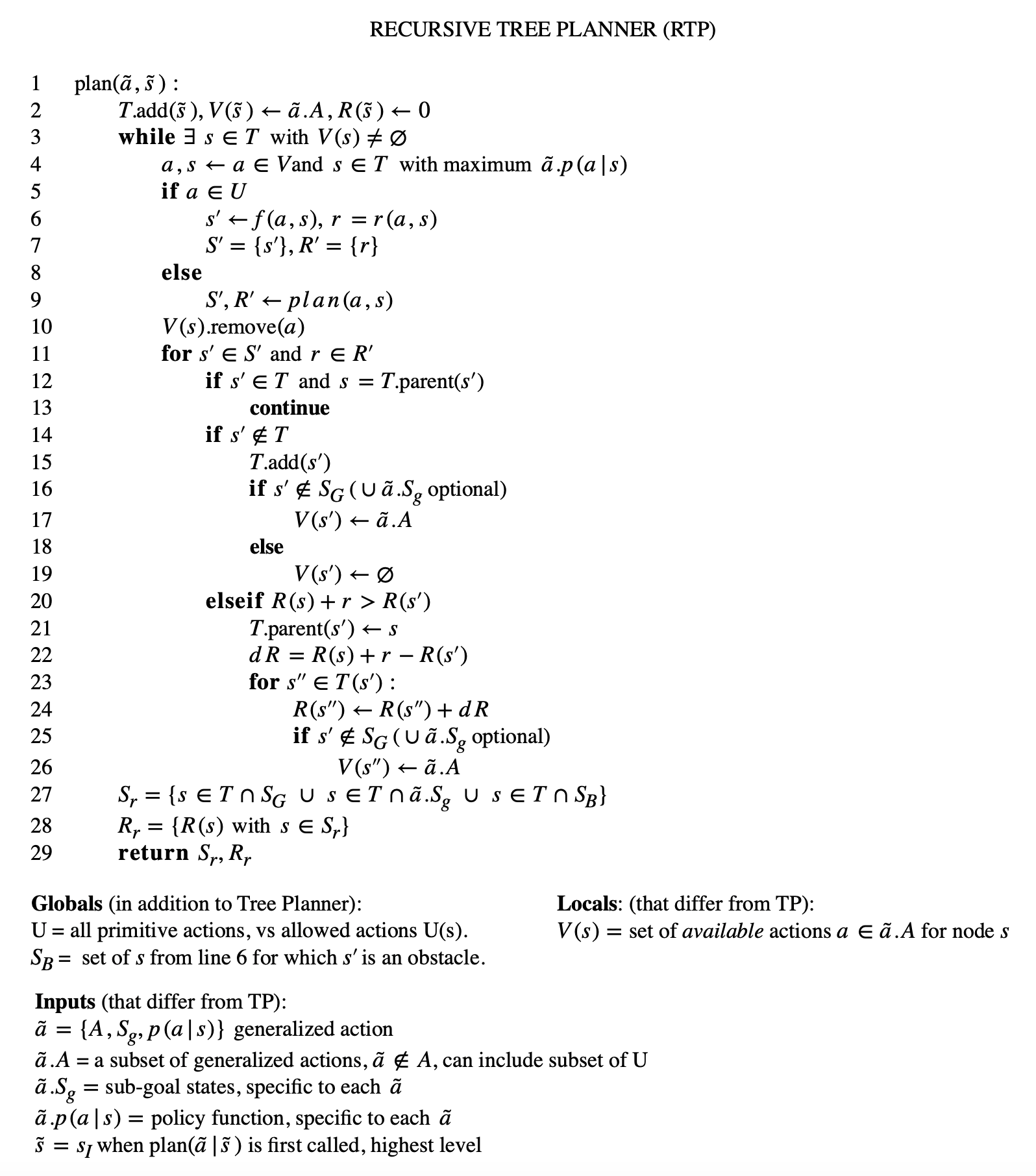} 
  \caption{Pseudocode for the hierarchical planner, Section~\ref{RTP}. Key differences from the TP are the recursive call to the planner itself, line 9, and the return of multiple states, line27, including subgoals in $\tilde{a}.S_g$ and boundary states $S_B$.}
\label{fig_RTP}
\end{figure}

\subsection{The Recursive Tree Planner (RTP)}
\label{RTP}

The single layer TP planner, above, is now modified to build the hierarchical RTP planner, Figure~\ref{fig_RTP}. The key change from the flat TP planner is that the RTP planner can call itself recursively. So, the planner can either take a primitive action step when $a \in U$, using the transition function $f(a,s)$, line 6, or it can take a \textit{generalized action} step, by calling the planner itself, line 9. The generalized action $a$ has all the information a planner or sub-planner needs to run, including any previously learned policies $p$. Hence, it replaces the policy $p$ argument in the TP planner. A generalized action, \ref{GA}, is to a planner or sub-planner, what a primitive action $u$ is to the transition function $f(u,s)$. To be clear, in Figure~\ref{fig_RTP}, $a$ denotes an action which can be either primitive or generalized.

Another change is that the RTP planner, line 9, can return a set of next states $S’$ in place of a single state $s’$. Therefore, the algorithm needs to check all of the states $s’ \in S’$ to see if any are already in the tree and, if so, replace the existing path to $s’$ when $R(s)+r>R(s’)$ in line 20 is satisfied.

The book keeping for the RTP, keeping track of when actions are available for expansion, can be more complicated than for the TP. This is because in the TP there is only one potential edge for each action taken from state $s$, so edges can be recorded as actions in $E(s)$. In the RTP, there can be multiple child states S’ coming from a single generalized action. One solution is to use a more complicated version of $E(s)$ which is what I do in practice for  efficiency. Instead, for brevity, the pseudo-code in Figure~\ref{fig_RTP}, explicitly checks to see if an edge already exists connecting $s$ to $s’$, line 12.

Another difference from the flat tree planner is that the tree can terminate at either a global goal state $s \in S_G$, or at a sub-goal $s \in \tilde{a}.S_g$, see lines 16 and 25. The termination at sub-goals is optional since one may want to return multiple sub-goals and some of these may be ancestors of other sub-goals so termination would cause some sub-goals to be missed. Returning \textit{multiple sub-goals} can speed up hierarchical tree search, as discussed for the inverted pendulum in \ref{exp_multi_subgoals}.

In addition, the planner can optionally return \textit{boundary states}, $S_B$, which are encountered when a search path attempts to transition into an obstacle state such as a wall. The $S_B$ are legal non-obstacle states and can include states a few steps back from the obstacle along the path. They are easy to find and store by adding $s$ to $S_B$ each time $s’$ in line 6 is a forbidden state. See more in Appendix~\ref{appendix_boundary_states}.

Although not shown, the RTP will return as many plans $\pi$ as states in $S_r$, line 29, calculated through backtracking. In practice, the backtracked $\pi$ are calculated and returned with $S_r$ at every level.

\subsection{Generalized Actions}
\label{GA}

A generalized action $a(g)$ contains all of the functions and parameters a planner or sub-planner needs in order to run, with the exception of the transfer function $f(u,s)$, rewards $r(u,s)$, and goal states $s_G$ for problem class $C$, which are made globally available to all levels of the RTP hierarchy. The initial state $s_I$ for each problem example is passed as input to the highest level planner. Thus, a generalized action is defined as

\begin{equation} 
\label{gen_a}
a(g) = \{ A_g, S_g, p_g(a|s),init_g(s), \gamma_g \}
\end{equation}

where 

\begin{itemize}[leftmargin=*]
\item $g$ parameterizes the sub-problem which the planner running $a(g)$ solves.
\item $A_g \in \tilde{A} \cup U$ is a set of actions which are available to be called by the planner running a(g), with $\tilde{A}$ the full set of generalized actions. To avoid loops,  $A_g$ cannot include a(g) itself or any higher level generalized action that calls $a(g)$.
\item $S_g$ is a set of sub-goal states for $a(g)$. Unlike $S_G$, see \ref{prob_form_1}, the $S_g$ need not be termination states and a sub-planner can return multiple sub-goals in $S_g$.
\item $p_g(a|s)$ is a stochastic policy with $a \in A_g$ and $s \in S$. It may have be previously learned from solutions to sub-problems (sp) which had $S_G(sp)=S_g$. It can alternatively be learned within the hierarchy from data generated while the planner solves some higher level problem.
\item $init_g(s)$ is optionally called at start of planning, line 2 in Figure~\ref{fig_RTP}, but not shown. Can be used to adjust the number of $A_g$, for example, to align with the number of objects in scene $s$.
\item $\gamma_g$ are planner parameters including maximum search tree size and depth.
\end{itemize}

Often the explicit parameter g will be dropped in favor of a dot notation:

\begin{equation} 
\label{gen_a2}
a=\{a.A, a.S_g, a.p(a|s), a.init(s), a.\gamma\}
\end{equation}

The goal parameter $g$ can directly index a sub-goal state $s_g \in S$. This is the case in the four rooms and inverted pendulum experiments in Section~\ref{experiments}. Alternatively, the goal index $g$ can refer to a sub-problem type, for example, $a(g=1)$ might be the generalized action which a robot uses to grab a 2d block as shown in Figure~\ref{exp_block_stacking}, and $a(g=2)$ might be the action which places one block on top of another. In addition, $g$ can index a specific object label $n$, so, for example, $a(g=grab,n)$ might be the generalized actions which grabs object $n$.

The goal parameter $g$ can also represent the \textit{location} of an object rather than an object label. This leads to the ability to use local scene properties to decide which object to act on next. It also can lead to approximate \textit{object number invariance}, as demonstrated in the experiments Section~\ref{exp_zero_shot} and discussed in \ref{implementation} and Appendix~\ref{appendix_number_invariance}. 

Object number invariance also relies on $a.init(s)$ to initialization the number of generalize actions $a.A_g$ to match the number of objects in the scene described by $s$.

In most cases, the best way to let the planner know that it has found a sub-goal is to define or learn a binary function which is true when $s \in S_g$. For many of the simple low-level generalized actions in the experiments below this function was learned along with policy learning, see Appendix~\ref{appendix_sub_goals}.

\subsection{Hierarchical Optimization}
\label{HO}

A multi-level RTP plan will recursively produce a primitive-action plan, which is of course the goal of the planner. This primitive-action plan, however, is not necessarily the optimal one that would in principle be found by a flat planner. That is because optimality is achieved only within each call to $plan(\tilde{a},\tilde{s})$ over the set of states it explores. Working up from the lowest level, I believe it is, however, recursively optimal \citep{dietterich}. 

One way to improve optimization at higher levels of the hierarchy is to increase the number sampled states available to search. This can, however, be expensive since each sample comes from a call to a lower level planner. To get more samples without extra calls to lower level planners, the RTP can return multiple sub-goals from those lower level planners. This can be far more efficient and boost planning performance as  demonstrated in Section~\ref{exp_multi_subgoals} for the inverted pendulum.

\subsection{Continuous Actions and States}
\label{continuous}

Adapting the RTP algorithm to continuous states requires finding a way to sample the continuous state space while maintaining reproducibility of paths in the search tree. Any path must consist of a series of actions which step through state space using the transition function. If states in the tree are approximated, for example by snapping to a grid, then there is no guarantee that paths in the tree will be consistent with the transition function. Therefore, any sampled states need to be precisely the ones found by following the transition function from other states in the tree. However, to keep the size of the search tree from getting too large, samples should also not be too close to each other.

To this end, I introduce the equals function $e(s,s’)$ in Formulation~\ref{prob_form_2}, which is true when states $s$ and $s’$ are within some minimal sampling distance. Like $r(u,s)$ and $f(u,s)$, the equals function $e(s,s’)$ is domain dependent but is a black box to the RTP. The equals function is used in lines 12 and 14 of Figure~\ref{fig_RTP} to decide whether a new state $s’$ is already in the tree $T: \exists \ e(s,s')=true, \ s \in T$ if  then $s’$ is in the tree. This spreads out the density of samples in $T$.

In experiments such as the lunar lander, Figure~\ref{exp_lunar_lander} and inverted pendulum Figure~\ref{exp_inverted_pendulum}, I have found that the equals function works very well and seems to require very little tuning. The minimum sampling distance just needs to be consistent with the typical step size taken by the transition function when taking a typical action.

For feasible plans — no rewards along the way — this is pretty much the end of the story. Most of the experiments in this paper require only feasible solutions, including the inverted pendulum which has continuous actions and states. For feasible plans, sub-tree swapping in lines 20-26 in Figure~\ref{fig_RTP}, is not needed. On the other hand, when the goal is to optimize the cumulative reward, lines 20-26 need to be modified and there is no longer a guarantee of optimality, although the algorithm does monotonically improve cumulative rewards $R(s)$ for nodes in the tree. Details are given in the Appendix~\ref{appendix_continuous_states}.

For continuous actions, I quantize and bin the actions. I have also tried using a random set of actions and that works too. For the experiments here at least, the solutions have been very insensitive to the number of bins.

\subsection{Zero-shot Transfer}
\label{zero_shot}

Perhaps the greatest benefit of a planner is the ability, in principle, to solve novel problems without any policy learning. In practice, at least some policy learning is often needed to make planning practical. What’s interesting, though, is that the policies learned for one class of problems, $C_i$, can often be exploited by the planner to solve other classes of problems, $C_j$.
I will refer to this as zero-shot transfer.

There are two primary mechanisms that the RTP uses to transfer learning from class $C_i$ to class $C_j$. The first is to use the greedy policies of $C_i$ as the starting point for the tree search for $C_j$, searching farther outside the greedy neighborhood as necessary. This has been referred to as \textit{near-greedy} search. It means that if the policies for $C_i$ get you somewhere sort-of close to a solution to $C_j$, it buys you a huge improvement in planning speed and accuracy.

The second mechanism is \textit{hierarchy}. In this case, policies for simper $C_i$ problems can be used as generalized actions by a higher level $C_j$ planner to sample the state space sparsely but efficiently. All that is needed from the $C_i$ here is to provide a reasonable sampling strategy and it is not uncommon to find $C_i$ which have this property.

I have found that the most powerful mechanism for zero-shot transfer is actually a combination of both near-greedy and hierarchy. There are two ways this gets implemented. First, the near-greedy search inside the low level planners running $C_i$ policies can tune those policies, in real time, to the $C_j$ problem. Second, the higher level planners for $C_j$ can call a combination of both $C_i$ generalized actions and primitive actions. In this case, if a $C_i$ sub-planner gets you close to a solution, the higher level planner can fine tune the solution at the higher level. In this sense, \textit{refinement} can happen at all levels and furthers the cause of zero-shot transfer.

In addition, $C_i$ policies can be designed to improve generalization through object number and background invariance, see \ref{implementation}, \ref{appendix_background_filtering}, \ref{appendix_number_invariance}. Such invariance allows $C_i$ policies to generalize to almost any scene and hence be used by a higher level policy to sample the state space. The tradeoff is that they don't \textit{see} obstacles so they crash into them. However, this can be exploited by 
having sub-planners return the boundary states $S_B$, \ref{appendix_boundary_states},  just before a crash, which helps the planner explore the contours of configuration space.

\section{Learning}
\label{learning}

\subsection{Learning Algorithm}
\label{learning_algorithm}

In the PLP framework of Section~\ref{PLP}, learning is a two step process: To start, the RTP generates a batch of plans $\{\pi\}$ for a set of problem examples. These plans include multiple sub-plans too, one for each unique generalized action in the RTP hierarchy. Each set of sub-plans is then used to learn or improve a policy for its corresponding generalized action. Note that the framework is iterative, so each time the RTP is run it uses whatever policies it has previously learned. The learning process is the same for plans and sub-plans, so I will show how it works for any batch of plans:

Each plan is a set of action-state pairs so each batch of plans is also a set of action-state pairs:

\begin{equation} 
\label{batch}
batch = \{ (a_0,s_0),(a_1,s_1),…(a_K,s_K) \}
\end{equation}

These are fed into a supervised learning algorithm which learns to predict the probability $p(a|s)$ of action $a$ from state $s$. The $p(a|s)$ is then used as the stochastic policy. The variable $a$ in $p(a|s)$ is actually an index into a set of generalized and primitive actions, but I will not make this explicit. I assume that the actions are mutually exclusive categorical variables so they can be converted to a one-hot representation, $O(a)$. A 1-hot variable, $O(a)$, has the property that for the subset of the data $(a_i,s_i)$ in (\ref{batch}) which satisfies $s_i=s_p$, for some particular $s_p$, the sum of $O(a_i)$ over the data will be proportional to the probability $P_{data}(a|s_p)$. Hence, to learn a functional approximation for $P_{data}$, I use the Cross Entropy loss function which learns one distribution from another:

\begin{equation} 
\label{cross_entropy}
E = -\sum_{i=0}^{K} O(a_i,s_i) \cdot log(p(a_i|s_i))
\end{equation}

For $p(a|s)$, I used a small neural network with a soft-max final layer which nicely complements the cross entropy loss function and forces $p(a|s)$ to sum to 1.0 over $a$ for any input $s$. To be clear, $s$ is the input to the neural network, and the number of outputs in the final layer is the number of possible actions $a$. The sum over $i$ is over the batch data in (\ref{batch}).

As an aside, the policies here are stochastic even for deterministic problems with discrete actions and states because there may be many equally good ways to solve the same problem. For example, in block stacking it may not matter which block is stacked next. Also, when the states are continuous, function approximation in $p(a|s)$ may not be able to distinguish between nearby states to which the planner assigns different actions.

\subsection{Gumbel-Softmax Alternative}
\label{gumbel}

Although its a bit off-track, I found that there was a completely different way to learn stochastic policies from 1-hot targets. The idea is to use a projector $M$ which is a matrix that converts the one hot vector $O(a)$ back to the action index: $a=M \cdot O$. Then the cross entropy loss function (\ref{cross_entropy}) can be replaced by a mean squared loss $E=(a-M \cdot p(a|s))^2$. In this case, it is better to think of $p(a|s)$ as a stochastic variable that directly generates the one-hot target $O(a)$. A natural choice to generate one-hot targets is the Gumbel-Softmax generator \citep{jang}, so I used it in place of the softmax function for the neural network output. The generator has a temperature variable that is slowly lowered during learning.

I tested this approach on block stacking, Figure~\ref{exp_block_stacking}, because there are many equally good ways to stack blocks so its policy is stochastic. The Gumbel-Softmax policy turned out to be exactly the same as the one learned using Cross Entropy.

\subsection{Implementation Considerations}
\label{implementation}

There are a number of implementation details for policy learning, many of which improve zero-shot transfer, Section~\ref{zero_shot}. These include background invariance, which allows policies learned in one scene to generalize to almost any other. For this, a lossy \textit{pre-filter} is used as a front-end to the policy, see Appendix~\ref{appendix_background_filtering}. This is especially useful for low level policies such as $grab$ and $place$ in Figure~\ref{exp_grab_place}. The resulting policies don’t see obstacles, but the planner can easily compensate by using near-greedy, hierarchical search, and boundary states $S_B$.

While pre-filters produce exact invariance, as planning problems become more complex, invariance can only be approximated. This can be achieved by using a 2d (or 3d) image-like input state representations and neural networks which limit their calculations to local features, such as a \textit{cnn}. This is complemented by using a 2d (or 3d) 1-hot representation for the actions $a$ in the policy p(a|s). For example, if an action is indexed by an object label $n$, then $n$ can be replaced by the position of the object $(x,y)$. The policy $p(n|s) \rightarrow p(x,y|s)$ then peaks at the location of the best object $n$ to act on, see Appendix~\ref{appendix_2d_1_hot_actions}. An example which uses 2d inputs and a 2d 1-hot representation for object number invariance is block stacking, where a policy learned for 3 blocks works perfectly for the much more difficult 5 blocks problem, \ref{speed_block_stacking}.

Finally, it is often the case that a high level planner has to choose not only what object to act on, but also how to act on it. For example, the planner may have to choose between the tasks $grab$ and $place$ and also decide which object to grab, a sub-task. Rather than learn one big policy which chooses both tasks and sub-tasks, policies can be split by task and sub-task and later combined using a generalized chain rule, as described in Appendix~\ref{appendix_policy_splitting}. This allows the sub-task policy, for example, to use 2d inputs and 2d 1-hot representations, while allowing the task policy to use a simpler 1d 1-hot representation. Neural networks can then be smaller, easier to learn, and generalized better.

\section{Experiments}
\label{experiments}

\subsection{Overview}
\label{exp_overview}

The experiments are designed to test how well the planner meets the goals of Section~\ref{design_goals}, and how the design properties in \ref{design_goals} help achieve those goals. Hence, the questions:

\begin{itemize}[leftmargin=*]
\item Does PLP policy learning improve RTP accuracy and speed? How well does PLP/RTP learn a greedy policy?
\item Does RTP hierarchy improve planner accuracy and speed? Do multiple sub-goals from sub-planners improve planner performance?
\item Is the RTP in the PLP framework good at zero-shot transfer? How do boundary states, hierarchy, near-greedy search, refinement at all levels, boundary states, $S_B$, and object number and background invariance all contribute to zero-shot?
\end{itemize}

In addition, can the RTP in the PLP framework learn multi-level policies simultaneously, and does the RTP generalize to continuous actions and states?

The experiments use either the Box2d physics simulator \citep{cato}, the Mujoco 3d physics simulator [E. Todorov et al], or toy world simulations in Python or Matlab. The neural networks are hand-written in C++ code and wrapped to run in Matlab or Python. They are very small with only 2 or 3 layers. All experiments were run on a MacBook pro and did not use the gpu. Hence solution times are useful only for comparison purposes.

Below, I will often use the shorthand $grab(n)$ to refer to generalized action $a(g=grab(n))$ and I will be sloppy about distinguishing between generalized actions $a$, their policies $a.p$, and the planner $plan(a,s)$ which runs generalized action $a$. 

Most of the experiments require only \textit{feasible} rather than optimal plans. Both feasible and optimal solutions are given for the four rooms problem, and terrain follow, is another example where optimal planning was tested.

A number of the problems use \textit{reward engineering} in the form of forbidden states to keep the allowed search space low, but this topic is left for the Appendix~\ref{appendix_reward_engineering}.

\subsection{Planning Speed and Accuracy as a Function of Policy Learning?}
\label{exp_speed_and_accuracy}

Table~\ref{table_speed} contains a number of examples showing that policies speed up planning and improve accuracy. With policies, performance is measured on test examples, not the training sets used to learn the policies.

\begin{table}
  \caption{Policy Speed and Accuracy}
  \label{table_speed}
  \centering
  \begin{tabular}{lllllll}
    \toprule
    Task &  Comment & Planner & L1/L2 Policy &Accuracy & Time (sec.)\\
    \midrule
    Place 3 boxes & Learn policy & Flat  & No/ - & 15/25  & 64\\
    Place 3 boxes & Uses 3 box policy & Flat  & Yes/- & 100\% & 0.4\\
    Place 9 boxes obstacles & Too slow, not solved & Flat  & No/- & 0\% & >1,212\\
    Place 9 boxes obstacles & Uses 3 box policy& Flat  & Yes/- & 100\% & 1.2\\
    \midrule
    Stack 3 boxes &  Slow & Flat & No/- & 100\% & 1874\\
    Stack 4 boxes &  Too slow, not solved & Flat & No/- & 0\% & >10,000\\
    Stack 5 boxes &  Too slow, not solved & Flat & No/- & 0\% & >>10,000\\
    Stack 3 boxes &  Learn policy & 2 Levels & Yes/No & 100\% & 23\\
    Stack 5 boxes &  Uses L1 policies & 2 Levels & Yes/No & 100\% & 623\\
    Stack 5 boxes &  Uses 3 box policy & 2 Levels & Yes/Yes & 100\% & 7.2\\
    \bottomrule
  \end{tabular}
\end{table}

\subsubsection{Place in box:}
\label{exp_place_in_box}
\begin{figure}
  \centering
  \includegraphics[width=\linewidth]{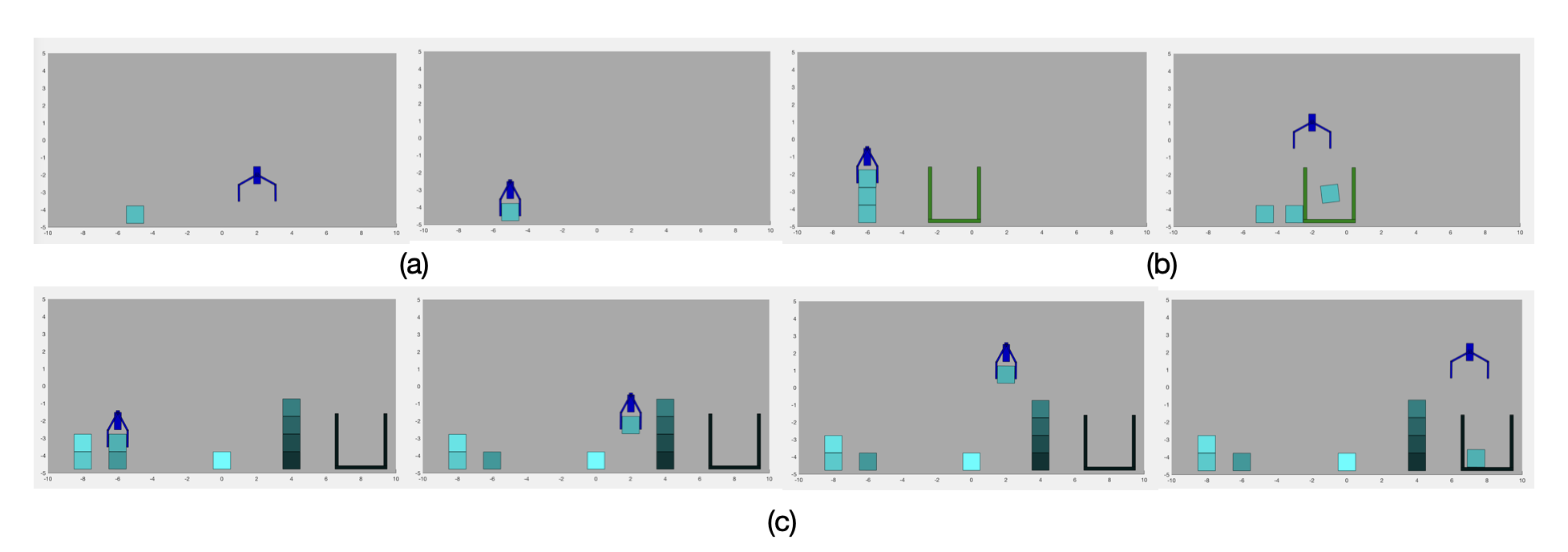}
  \caption{Box2d robot problems. (a) Simple $grab$, learned from scenes with just one object. (b) Simple $place$ in box learned from scenes with 1-3 objects (c) $place$ with obstacles and many boxes can be solve easily by planner using the 1-3 box policy from (b) by performing a near greedy search.}
  \label{exp_grab_place}
\end{figure}

The first test is the simple place-in-box problem with 3 boxes shown in Figure~\ref{exp_grab_place}b. A pure flat (non-hierarchical) planner needs an average of 64 seconds to solve this problem and only succeeds 15 out of 25 times. After using these 15 successful plans to learn a policy, the planner with policy has 100\% accuracy and the speed goes down to 0.4 seconds per example.

Next, the 3 box policy is used to solve the same problem, but with 9 boxes and with obstacles, Figure~\ref{exp_grab_place}c. With no policy, this problem took too long to solve because, with 9 boxes, there are too many possible states to explore. However, using the 3-box-no-obstacle policy, above, the problem was solved in 1.2 seconds. Note that this is also an example of zero-shot transfer to the extent that 9 boxes with obstacles is a different problem class.

\subsubsection{Block stacking:}
\label{speed_block_stacking}
\begin{figure}
  \centering
  \includegraphics[width=\linewidth]{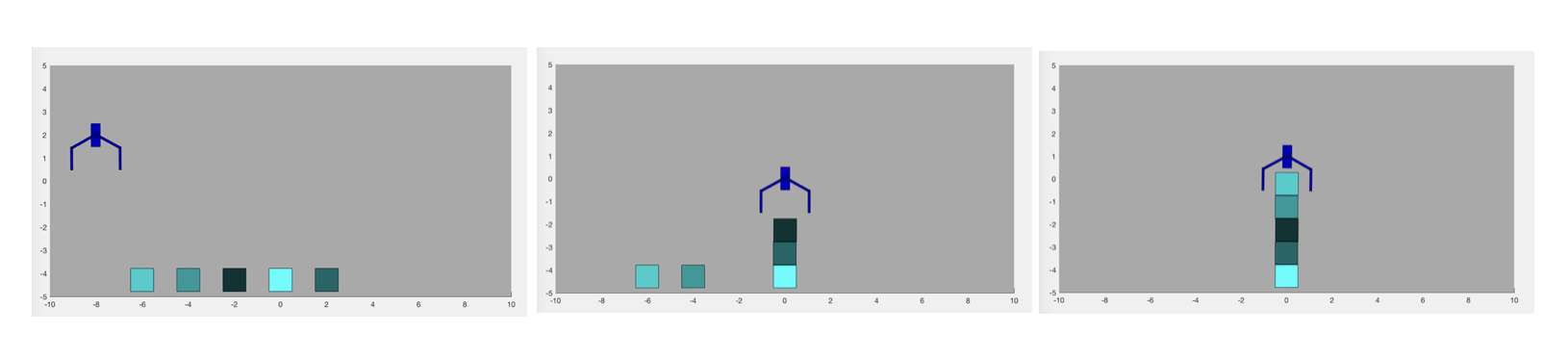}
  \caption{Block stacking. A flat 5 box plan using primitive actions takes too long to find. Using generalized actions $grab(n)$ and $place\_on(m)$ with their learned policies, a level 2 planner with no policy at level 2 easily finds a solution. When a level 2 policy is learned for the 3 block problem, it generalizes to the 5 block problem, speeding up planning, see Table~\ref{table_speed}}
  \label{exp_block_stacking}
\end{figure}

Block stacking, Figure~\ref{exp_block_stacking} is a long horizon problem, typically requiring about 80 primitive actions. Therefore, a pure flat planner is not able to solve the 4 or 5 block problem in reasonable time, while the 3 block stack took a slow 1,874 seconds. Therefore, a 2-level hierarchical planner was built which uses two level 1 generalized actions: $grab(n)$ and $place\_on(m)$. $grab(n)$ has a policy which learned to grab object $n$, while $place\_on(m)$ has a policy which learned to place the object-in-the-gripper onto object $m$. Using these in a level 2 planner reduces the 3 block stacking time from 1,874 seconds down to 23 seconds, while 5 block stacking, which was not feasible before, can now be done in 623 seconds.

To further speed up block stacking, a level 2 policy was learned for the 3-block level 2 problem. This policy was then applied to the much more difficult 5-block problem without any modification to the policy. This was easily implemented using generalized action initialization \ref{GA}, combined with local policy functions, \ref{implementation} and Appendix~\ref{appendix_2d_1_hot_actions}. It reduced the time to stack five blocks down from 623 seconds to 7.2 seconds.

\subsubsection{Inverted pendulum:}
\label{exp_inv_pendulum}
\begin{figure}
  \centering
  \includegraphics[width=\linewidth]{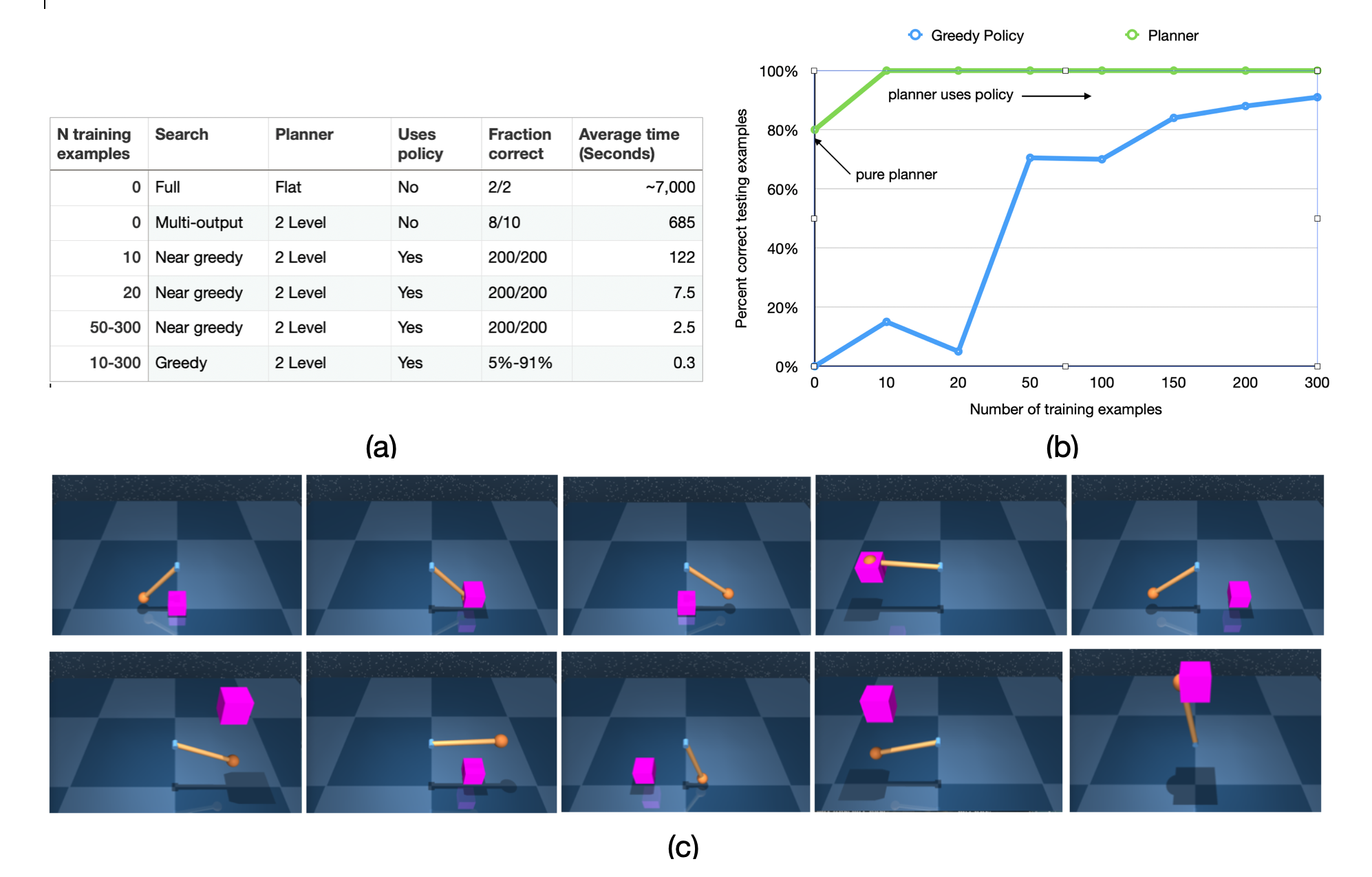}
  \caption{Inverted Pendulum in MuJoCo solved using a 2 level RTP planner. Policies are learned simultaneously for both levels. (a) Improvements in planner speed and accuracy as policies are improved with more examples. (b) Accuracy for the planner-with-policies and the pure greedy policies as a function of the number of training examples. (c) An example solution, showing the level 1 sub-goals as purple boxes. This solution requires swinging back and forth to gain speed and requires 187 primitive steps.}
  \label{exp_inverted_pendulum}
\end{figure}

Finally, consider the inverted pendulum problem, \citep{levy} shown in Figure~\ref{exp_inverted_pendulum}. For this problem, a 2-level planner simultaneously learned policies for both levels, and it is interesting to see how planning accuracy and speed improves with the number of examples used for learning, as shown in Figure~\ref{exp_inverted_pendulum}a. With no policies, the 2-level planner solves 8/10 problems correctly taking an average of 685 seconds per example. Once the 8 correct examples are used to learn policies, the planning accuracy goes up to 100\%, and planning speed goes down to 122 seconds. After 300 examples, speed is down to 2.5 seconds.

\subsection{Greedy Policy Accuracy as a Function of the Number of Examples?}
\label{exp_greedy_accuracy}

For the inverted pendulum, the reason accuracy goes up almost immediately in Figure~\ref{exp_inverted_pendulum}a,b, but speed goes down slowly is that the planner tries the greedy policy first and then searches near it and then farther away. The crude initial policy improves planning enough to solve the problem in reasonable time, but the search needs to explore pretty far from greedy. At 300 examples, the greedy policy has improved so much that the planner can find a solution very close to the greedy one. In fact, after 300 examples, the greedy policy alone without any planning is up to 91\% accuracy and takes only 0.3 seconds. This compares well with the hierarchical reinforcement learning approach in \citep{levy} from which the inverted pendulum problem was taken. The learning curve in Figure~\ref{exp_inverted_pendulum}b is similar to the one in \citep{levy}.

So this is an example where the PLP framework can be used to learn a stand alone policy which can then be divorced from the planner. In this case, the PLP framework is used as an alternative to traditional reinforcement learning (RL) methods, although it does require knowing the dynamics, while RL typically does not. Once outside the planner, this greedy policy is not as robust as the planner-with-policy, but it may be used, for example, to initialize other RL algorithms in a real-world environment with noisy or unknown dynamics.

In many of the other experiments here, such as in block stacking, the learned policies do work well at the greedy extreme without the planner.

\subsection{Hierarchical vs Flat Planning Speed ?}
\label{exp_hierarchical}

\begin{figure}
  \centering
  \includegraphics[width=\linewidth]{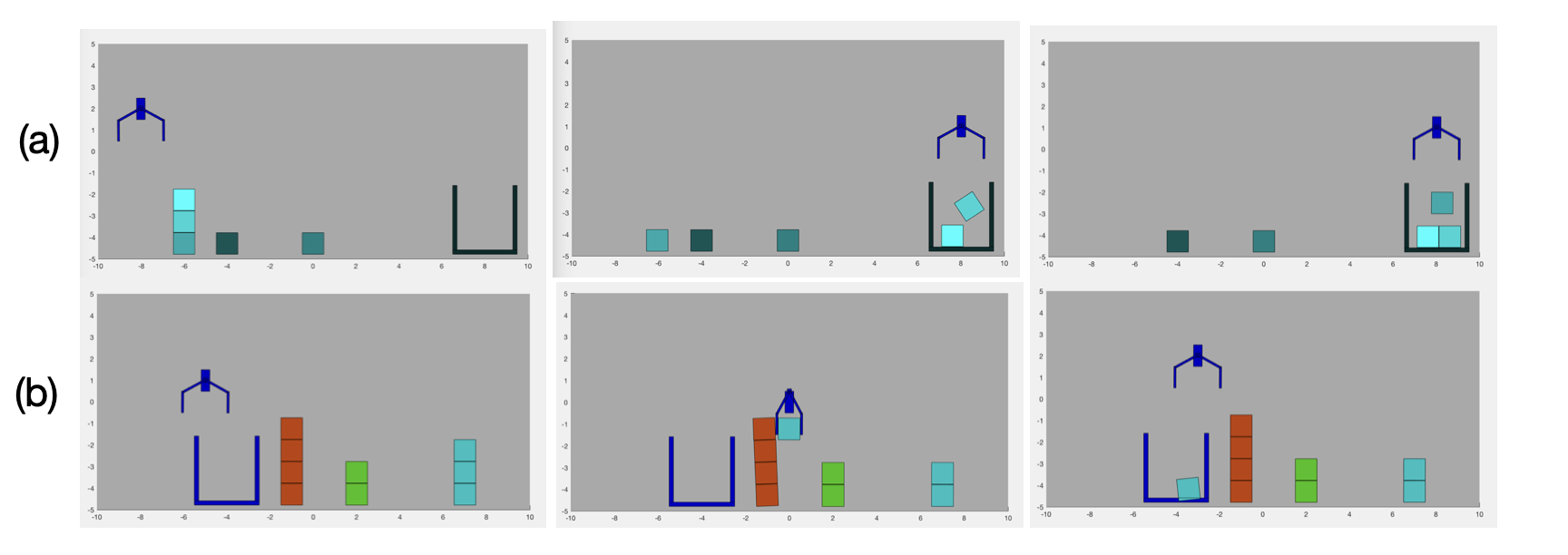}
  \caption{(a) Covered pick \& place uses grab and place to remove top two blocks to get to target block. (b) Planner solves problem with obstacles without prior knowledge of obstacles.}
  \label{exp_pick_place}
\end{figure}

The experiments in Table~\ref{table_hierarchy} demonstrate that hierarchy can produce a huge improvement in planning speed and can make planning possible. When the planning horizon is large, it can take too long to search through primitive action paths to find a problem solution. This is the case for pick \& place when the target object is covered by other objects as shown in Figure~\ref{exp_pick_place}a. On the other hand, the level 2 planner using generalized actions $grab$ and $place$, but without further learning, was able to find a solution in only 73 seconds, even though the solution takes 97 primitive actions. A flat planner for this problem was not able to solve it in reasonable time. Adding obstacles, Figure~\ref{exp_pick_place}b, to pick \& place also failed at level 1, but was easily solved in 2.2 seconds using a level 2 planner. The block stacking, Figure~\ref{exp_block_stacking}, horizon is just too long for 4 and 5 blocks to be solved by a flat planner, but is solved in 194 and 623 seconds respectively using a level 2 planner.

\begin{table}
  \caption{Hierarchical vs Flat Planning}
  \label{table_hierarchy}
  \centering
  \begin{tabular}{lllllll}
    \toprule
    Task &  Comment & Planner & Horizon & Accuracy & Time (sec.)\\
    \midrule
    Pick \& place covered & primitive actions & Flat  & n/a &  0\%  & >10,000\\
    Pick \& place covered & grab, place & 2 Level & 97  &  100\% & 73\\
    \midrule
    Pick \& place w obstacles & primitive actions & Flat  & n/a &  0\% & >1,600\\
    Pick \& place w obstacles & grab, place & 2 Level & 22 &  100\% & 2.2\\
    \midrule
    Stack 3 boxes &  primitive actions & Flat      & 25 & 100\%  & 1874\\
    Stack 3 boxes &  grab, place\_on  & 2 Level & 25 &  100\% & 23\\
    Stack 4 boxes &  primitive actions & Flat      & n/a & 0\%     & >10,000\\
    Stack 4 boxes &  grab, place\_on  & 2 Level & 58 & 100\%  & 194\\
    \midrule    
    Inverted pendulum &  primitive actions & Flat & 67-274 & 2/2 & $\sim$7,000\\
    Inverted pendulum &  multiple sub-goals & 2 Levels & 67-274 & 8/10  & 685\\
    \bottomrule
  \end{tabular}
\end{table}

\subsubsection{Simultaneous multiple sub-goals}
\label{exp_multi_subgoals}

One way in which the RTP can exploit hierarchy is to search for many sub-goals $s_g \in S_g$ simultaneously. This is one way that the hierarchical planner can be much faster than a flat planner, even when run as a pure planner without any learned policies. This is seen in Table~\ref{table_hierarchy} for the inverted pendulum where the flat planner takes about 7,000 seconds to plan, while a hierarchical planner returning multiple sub-goals is over 10 times faster. Again, this is true even without any policy learning. 

The reason for the large speedup for the inverted pendulum is that many sub-goal paths share sub-paths in the same search tree. Thus, using different search trees to find different sub-goals repeats many of the same calculations multiple times. On the other hand, not all problem classes will benefit from this speedup, since sub-goals may not necessarily share sub-paths.

\begin{table}
  \caption{Zero-Shot Transfer}
  \label{table_zero_shot}
  \centering
  \begin{tabular}{lllllll}
    \toprule
    Training Class(es) $C_i$ & Test Class $C_j$ &   $C_i/C_j$ Levels & $C_i/C_j$ Figures \\
    \midrule
    place 1-3 boxes  		&  place 9 boxes, obstacles 			&1/1 		& ~\ref{exp_grab_place}b/c 						\\
    grab, place         		&  pick \& place, target covered		 	&1/2 		& ~\ref{exp_grab_place}/~\ref{exp_pick_place}a 		\\
    \midrule
    grab, place         		&  pick \& place, obstacles  		 	&1/2 		& ~\ref{exp_grab_place}/~\ref{exp_pick_place}b		\\
    grab, place         		&  pick \& place, touching, friction 		&1/2 		& ~\ref{exp_grab_place}/~\ref{exp_friction}b 			\\
    \midrule
    grab, place\_on 		&   stack 2-5 boxes     	 			&1/2 		& /~\ref{exp_block_stacking} 						\\
    grab, avoid\_left/right 	&   place on, extreme re-use 		 	&1/2 		& ~\ref{exp_avoid_place}a/~\ref{exp_avoid_place}b 		\\
    stack 3 blocks   		&   stack 5 blocks, much harder      		&2/2 		& /~\ref{exp_block_stacking}  						\\
    short controlled flight 	&   land on moon or hover 	 		&1/2 		& ~\ref{exp_lunar_lander}a/~\ref{exp_lunar_lander}b 	\\
    \midrule    
    four rooms original barrier &  four rooms, different barriers 	 	&3/3 		& ~\ref{exp_four_rooms}d/~\ref{exp_four_rooms}e,f 		\\
    \bottomrule
  \end{tabular}
\end{table}

\subsection{Zero-Shot Transfer}
\label{exp_zero_shot}

As discussed in Section~\ref{zero_shot}, the RTP uses near-greedy search, hierarchy, and refinement at all levels to exploit policies learned for class(es) $C_i$ to speed up or make feasible solutions to some other class(es) $C_j$. Additional help comes by using boundary states $S_B$, and policies which are at least approximately object number and background invariant, Section~\ref{implementation} and Appendix~\ref{appendix_background_filtering},\ref{appendix_number_invariance}.

\subsubsection{Pick \& place}
\label{exp_zs_pick_and_place}

Table~\ref{table_zero_shot} lists the experiments here which demonstrate zero-shot transfer. In the first row, $C_i$, is the $place$ task in Figure~\ref{exp_grab_place}b which was learned from examples with 1 to 3 objects with no obstacles. Without any further learning, the $C_i$ policy solves the 9 object $C_j$ problem with obstacles in just 3.8 seconds. For this problem with obstacles, crashing into the obstacle is a forbidden state, like a wall, so actions which crash into the wall fail. The planner tries the greedy policy for the simple $C_i $ $place$, which is background invariant, Appendix~\ref{appendix_background_filtering}, so it goes crashing into the obstacle as though it were not there. The planner then quickly finds a near-greedy policy which takes it over the obstacle, see Figure~\ref{exp_grab_place}c. The plan ends up being the pure greedy $C_i $ up to the obstacle, a few non-greedy actions to go over, then returns to pure greedy to place the object in the box.

The next three experiments — next three rows — all use the two simple $C_i$ classes, $grab(n)$ and $place()$, shown in Figure~\ref{exp_grab_place}a,b. $grab(n)$ was learned from problem examples with only one block, while $place()$ was learned from examples with 1 to 3. These level 1 policies are then used as generalized actions by a level 2 hierarchical planner to solve pick \& place when the target object is covered, Figure~\ref{exp_pick_place}a. The planner discovers that it can repeatedly use grab and place to get non-target objects out of the way. This is a brand new use of grab and place and was never hinted at when they were learned. As an aside, it is also an example where the planner performs existential quantification as in “does there exist an object p which can be moved out of the way in order to be able to grab object q?”

The next $C_j $ problem, Figure~\ref{exp_pick_place}b, is pick \& place with obstacles. What is interesting here is that the level 2 hierarchical planner has two options: first, is to use a near-greedy solution inside the call to the level 1 $place$ planner, as in Figure~\ref{exp_grab_place}c. In this case, the hard work is done inside the level 1 planner, allowing the level 2 planner to call $grab$ and $place$ as though there was no obstacle. But there is second option: the level 1 planners can be restricted to run only greedy policies. Then when level 1 $place$ crashes into the object it returns the boundary state $s_B$ just before the crash. If the set of available actions $a.A$ for the level 2 planner includes the set of primitive actions $u \in U$, then the level 2 planner can call a few primitive actions $u$ to get around the object and then call $place$ again to finish up. So the planner can refine policies at either level.

\begin{figure}
  \centering
  \includegraphics[width=\linewidth]{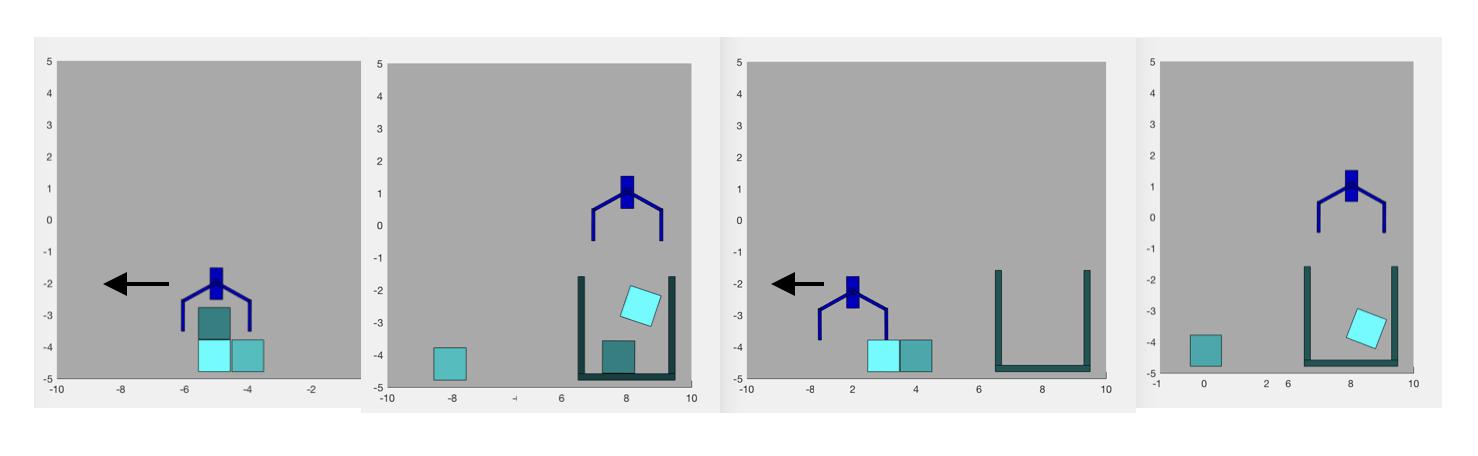}
  \caption{Pick and place, but gripper is blocked by object up against target (lightest) object. Left scene, planner discovers that friction moves stack with target underneath using open gripper. Right scene, planner discovers that friction of gripper tip moves target to make room for gripper. In both scenes $grab$ and $place$ used by the planner in novel and unanticipated ways.}
  \label{exp_friction}
\end{figure}

The third pick \& place $C_j$ problem is shown in Figure~\ref{exp_friction}. Here, the twist is that the gripper can’t actually grab the target object because other objects are too close and obstruct the gripper. Even though the $C_i$ $grab$ and $place$ have no clue that objects can be moved using friction, the level 2 planner discovers that it can exploit a few primitive actions $u$ to use friction to move objects. Two different strategies are discovered, as shown, for the two different $C_j$ problems.

\subsubsection{Place on stack}
\label{exp_zs_place_on}

\begin{figure}
  \centering
  \includegraphics[width=\linewidth]{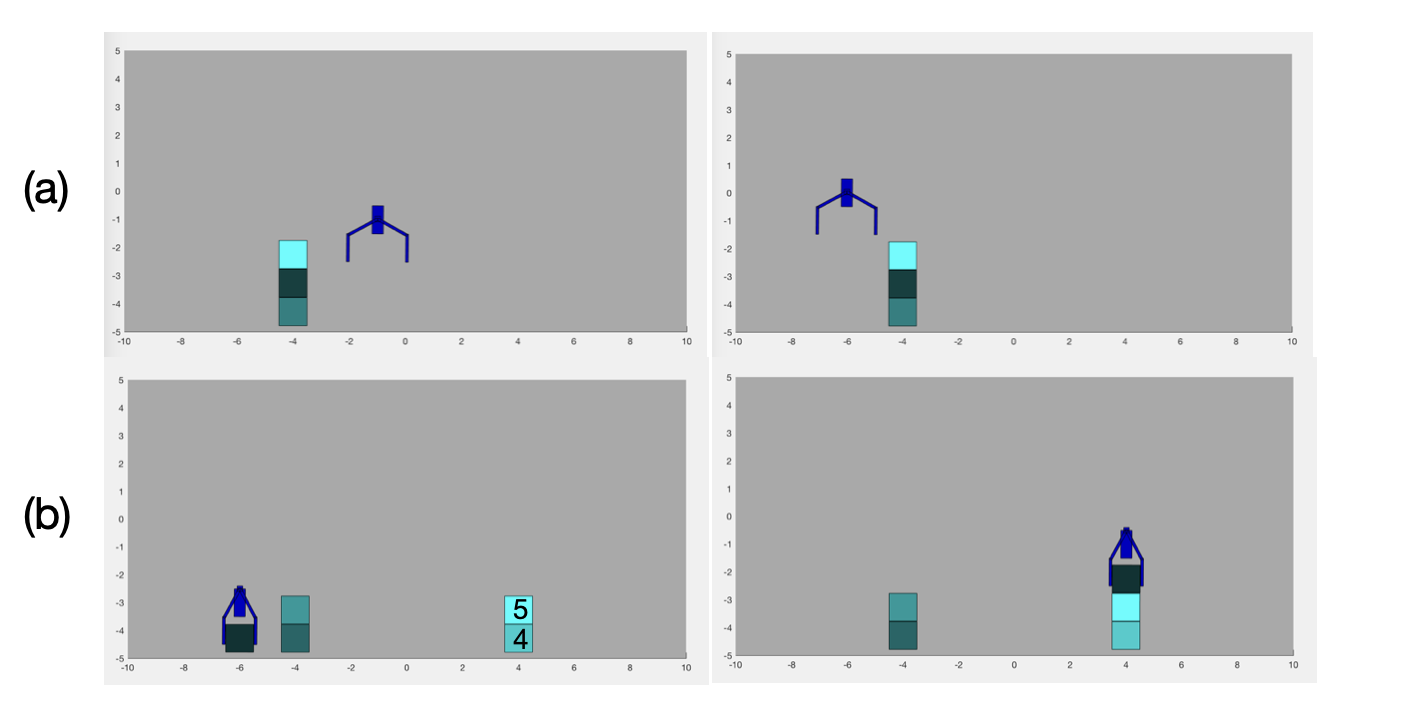}
  \caption{(a) The level 1 task $avoid\_left$ learns to avoid an obstacle to the left of the gripper (b) At level 2, $avoid\_right$ together with a version of $grab$ are used in unanticipated ways to avoid the obstacle and place the object in the gripper onto the stack to the right.}
  \label{exp_avoid_place}
\end{figure}

Block stacking, discussed in \ref{exp_speed_and_accuracy} and \ref{exp_hierarchical}, is another example where the hierarchical planner uses simple $C_i$ policies, $grab$ and $place\_on$, to solve the difficult 4 and 5 block $C_j$ problems. It may not be surprising that $grab$ and $place\_on$ work as sub-planners for this problem since they are clearly sub-problems. On the other hand, the next experiment is designed to test whether a hierarchical planner can use arbitrary low level $C_i$ policies to solve high level problems. The $C_j$ problem, Figure~\ref{exp_avoid_place}b, is to place the object in the gripper onto the stack to the right while avoiding the obstacle. To do so, however, a level 2 planner is given level $C_i$ policies which are not sub-problems. The first is $grab(n)$ even though no grabbing is required, and the other two are $avoid_{left}(m)$, Figure~\ref{exp_avoid_place}a, and $avoid_{right}(m)$, which move the gripper up and around a stack of objects with top object $m$.

The solution in Figure~\ref{exp_avoid_place}b surprised me: First, $avoid_{right}(1)$ was used to get object 1 high up so it won’t crash into the obstacle. However, object 1 is not the top object on the obstacle, but is the object in the gripper, so the object in the gripper is avoiding itself! The level 2 planner then uses $grab(4)$ to try to grab the object underneath object 5, which has the effect of placing the object in the gripper \textit{onto} object 5. Again, this is a highly non-standard and unanticipated use of $grab(n)$.

\subsubsection{Four rooms}
\label{exp_zs_four_rooms}

\begin{figure}
  \centering
  \includegraphics[width=\linewidth]{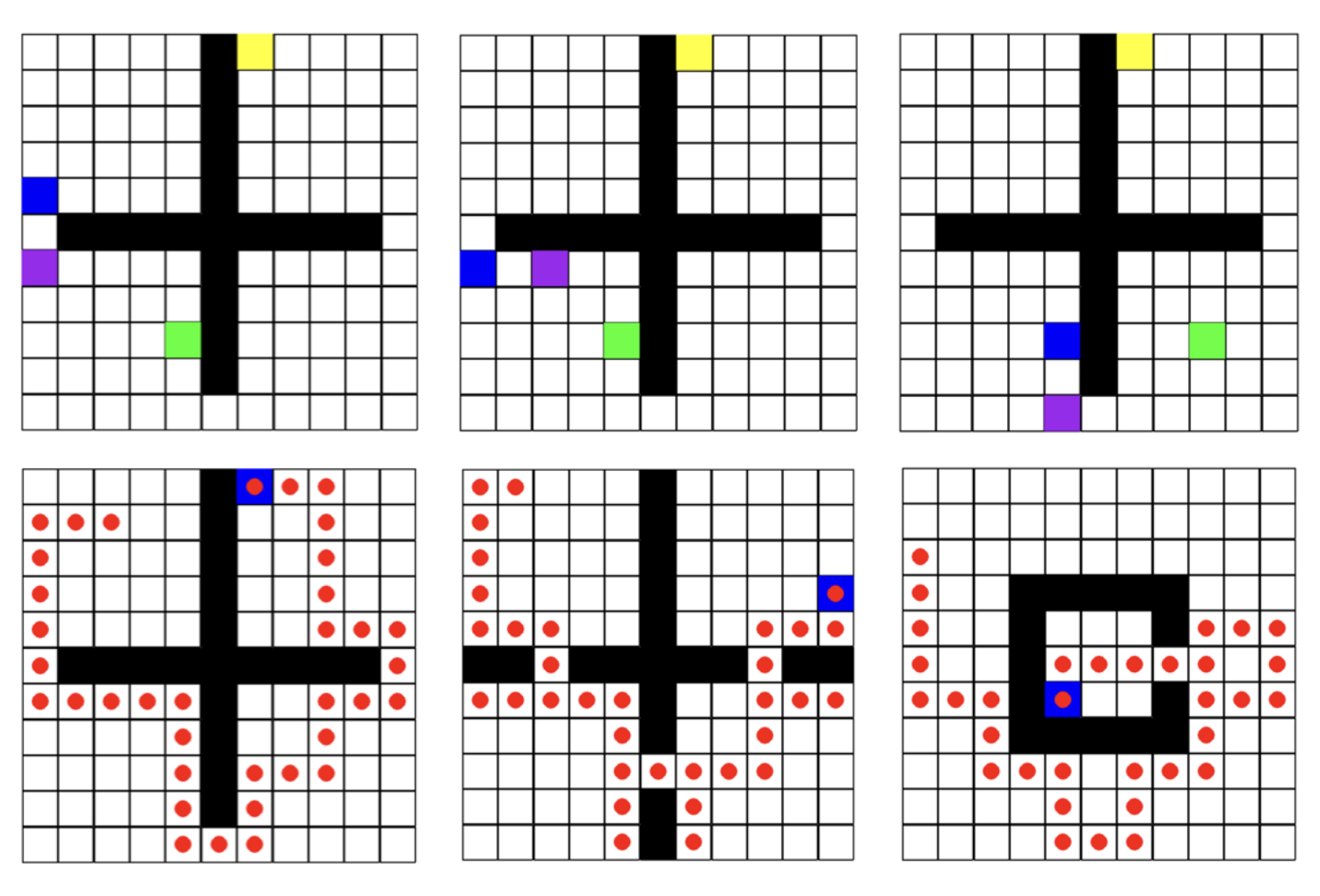}
  \caption{Four rooms. Goal is to get blue block to yellow block. All policies for a 3 level planner are learned simultaneously. Top row shows sub-goals: purple for the lowest level and green for level 2. Below, feasible solution on left. Policies learned for left barrier find zero-shot feasible solutions for right two barriers. Optimal solutions are found too, see Figure~\ref{exp_four_rooms2}}
  \label{exp_four_rooms}
\end{figure}

The four rooms problem \citep{levy}, Figure~\ref{exp_four_rooms}, is used below to demonstrate that the RTP planner can be used for simultaneous multi-level policy learning. It also can be used to demonstrate that near-greedy search at multiple levels simultaneously can solve novel problems. First, 3-levels of policies were learned to solve the problem with the barrier shown in the first four panels. Those policies were then used without further learning to solve the next two panels with different barriers. I have shown feasible solutions for the problem (not shortest paths) to give a sense of how the planner searches for novel solutions. As can be seen, the planner tries the greedy path for the original barrier, but when it fails it modifies that path a little to find a similar path to the new barrier. Optimal paths can be found too, Figure~\ref{exp_four_rooms2}, as discussed in \ref{exp_opt_four_rooms}.

\subsubsection{Lunar lander}
\label{exp_zs_lunar_lander}

\begin{figure}
  \centering
  \includegraphics[width=\linewidth]{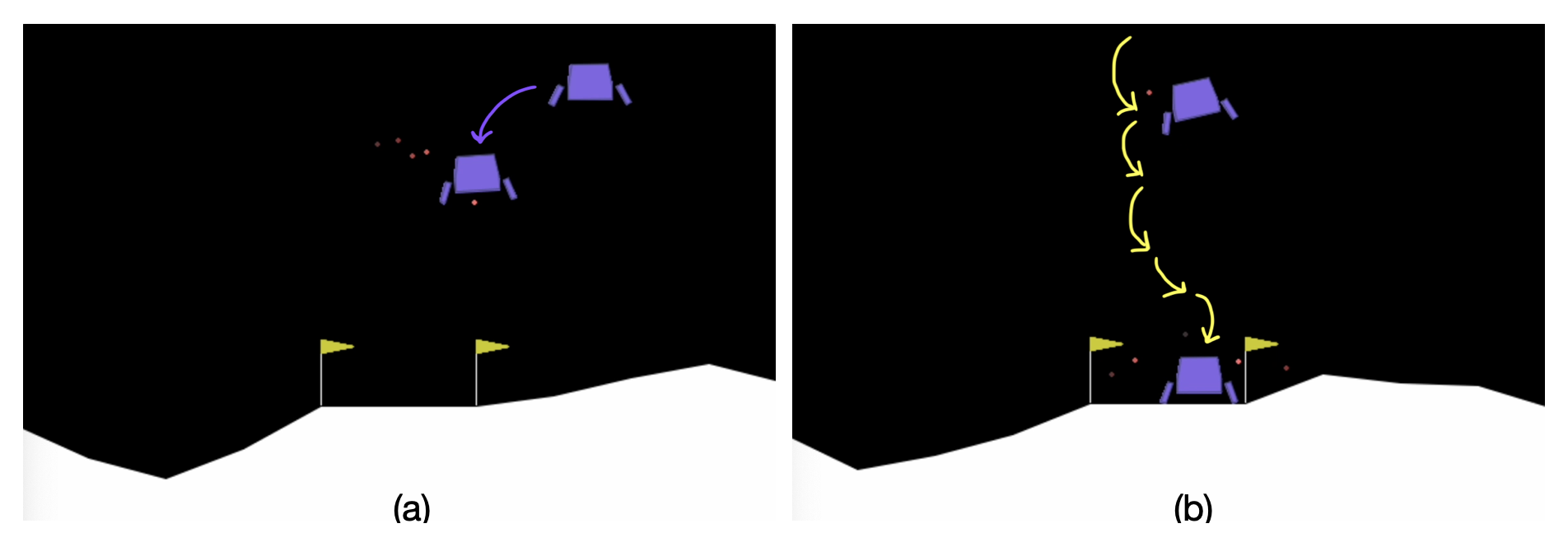}
  \caption{Lunar Lander. (a) A policy was learned for short controlled flight using the same rewards as the full problem but with a low threshold. (b) Policy in (a) is used as a generalized action by a level 2 planner to land on moon.}
  \label{exp_lunar_lander}
\end{figure}

The classic lunar lander benchmark from the Gymnasium test suite \citep{farama} is a long horizon problem which is difficult for a flat planner to solve, see Figure~\ref{exp_lunar_lander}. The goal is a fast, smooth, controlled flight and landing on the moon. This turns out to be another example where solving a much simpler problem $C_i$ by learning a level 1 policy allows a level 2 planner, without further learning, to solve the full lander problem $C_j$. But most importantly, the simpler problem knows nothing about landing on the moon, so it can be used to solve other problems such as how to hover at some altitude above the moon!

The lunar lander problem does not seem to be naturally hierarchical. To my surprise, both tree search and hierarchy worked quite well in this context. Because there is no obvious hierarchy, the first job was to find a relaxed sub-problem $C_i $that could be realistically solved and used as a level 1 generalized action. For this relaxed problem, I had the lander fly a very short distance while requiring a small improvement in flight control, Figure~\ref{exp_lunar_lander}a. This was implemented by finding solutions, from arbitrary initial states, that have a cumulative reward greater than a small threshold. Since positive rewards are obtained by flying well, this produces nicely controlled flights over short distances. The rewards and cumulative rewards are exactly those provided for the full problem. 

A random tree search, mentioned in section \ref{TP}, was used to find solutions from about 20 examples of the short flight problem, and a small three layer neural network policy was learned. Since the state space is continuous, the techniques in Section \ref{continuous} were used for sampling. 

A level 2 planner then used this short flight policy as a generalized action. Using a pure planner at level 2, it was able to find, for example, the solution in Figure~\ref{exp_lunar_lander}b, with cumulative reward 267 (above the 200 threshold in \citep{farama}) using a small 59 state search tree. In addition to using the learned short-flight policy, the level 1 planner returned near-crash boundary states just above the moon surface which could be used by the level 2 planner. Landing on the moon takes about 200 primitive actions which would take a flat planner a very long time to discover. As mentioned, the same system finds hovering solutions as well.

\subsection{Learning Multi-level Policies Simultaneously}
\label{exp_multi_learning}

To test of the ability of the RTP algorithm to learn policies at all levels of the hierarchy simultaneously, I used the RTP to implement the goal-conditioned hierarchy described in \citep{levy}. In this framework, the variable $g$ in generalized action $a(g)$ directly indexes a particular sub-goal state $s_g \in S_g$. (This is but one of many ways index $g$ can be used, Section \ref{GA}) Therefore, by choosing to take action $a(g)$, a higher level planner, is choosing to run a sub-planner $plan(a(g),s)$ which attempts to find a path to the sub-goal state $s_g$.

\subsubsection{Four rooms}
\label{exp_multi_four_rooms}

To be concrete, consider the four room problem shown in Figure~\ref{exp_four_rooms} with a three level hierarchy as described in \citep{levy}. In this problem, the goal state for the highest level planner $a(g_2)$ is the yellow block and the highest level goal $s_G$ is achieved by moving the blue block, using primitive actions $u$, to the yellow block, while avoiding obstructions. Planner 
$a(g_2)$ has its own set of generalized actions $a(g_1) \in a(g_2).A$ with each $g_1$ an index into a sub-goal state which is shown in green. This means that planner $a(g_1)$ needs to move the blue  block to the next green block $s(g_1)$ as a sub-problem. In turn, each $a(g_1)$ has a set of generalized actions $a(g_0) \in a(g_1).A$ with each $g_0$ an index into a sub-goal state shown in purple. The example in Figure~\ref{exp_four_rooms}, shows some intermediate steps in the top row, as well as the complete learned path in the left hand panel of the second row.

The full three-level RTP with the above choices was run starting from scratch with no policies at all levels. A batch of 10 correctly solved training examples produced initial plans at all levels which were fed into neural networks to learn policies for each of the three levels. Hence, all three levels were learned simultaneously, as in \citep{levy}. For this simple problem, only one iteration of planning/exploration followed by learning was necessary. After learning, 100\% of test examples were solved in average time 0.024 seconds, which is about 10 times faster than the average time of 0.27 seconds needed by a flat planner without policies.

Comparing to \citep{levy}, the approach here required roughly the same number of training examples, maybe one or two more, but I didn’t try fewer than 10. The supervised learning algorithms used here are simpler and potentially more robust than the reinforcement learning algorithms in \citep{levy}. However, the approach here may not scale as well, and it does require a dynamics simulator which \citep{levy} in principle does not.

\subsubsection{Inverted pendulum}
\label{exp_multi_inverted_pendulum}

In Sections~\ref{exp_speed_and_accuracy}-\ref{exp_hierarchical}, I looked at policy learning and the benefits of hierarchy in the inverted pendulum problem from \citep{levy}. The primary motivation for trying this problem, however, was to test simultaneous multi-level policy learning using the RTP. As shown in Figure~\ref{exp_inverted_pendulum}c, a two level hierarchy was implemented with the low level subgoal getting to the next purple box, while the high level chooses this sub-goal by choosing a generalized action parameterized by purple box location. Simultaneous multi-level policy learning succeeds here: after solving just 8/10 examples without policies, the planner learned policies which increased accuracy to 100\% and increased speed by a factor of 5. Then through PLP bootstrapping, Section~\ref{PLP}, speed steadily increased as more examples were solved.

A typical solution is shown in Figure~\ref{exp_inverted_pendulum}c where the pendulum needs to swing back and forth to gain enough speed to reach the top. What makes this problem difficult is the number of primitive steps needed which is 187 here. This is why it benefits so much from a hierarchical solution.

Comparing the result here to \citep{levy}, the RTP greedy policy accuracy, Figure~\ref{exp_inverted_pendulum}b, is about the same or a little better, as a function of the number of training examples. The big difference is between pure greedy policies and the near-greedy planner, with the planner accuracy at 100\% after learning from just 8/10 examples.

On a different note, the inverted pendulum might not look like a good fit for tree search since both actions and states are continuous. However, action binning and sampling with an equals function, Section~\ref{continuous}, seems to work well. I quantized the single joint action into 21 bins, a choice made arbitrarily before any experimentation. The equals function was set to return false if the pendulum joint angle or joint velocity was less than 1/100 of their respective ranges. I later tightened this to 1/150, but this required no change to any learned policies or any algorithm. It should be noted that since the states on the tree remain continuous, all policies take continuous states as input.

\subsection{Optimal Paths}
\label{exp_optimal_paths}

\begin{figure}
  \centering
  \includegraphics[width=\linewidth]{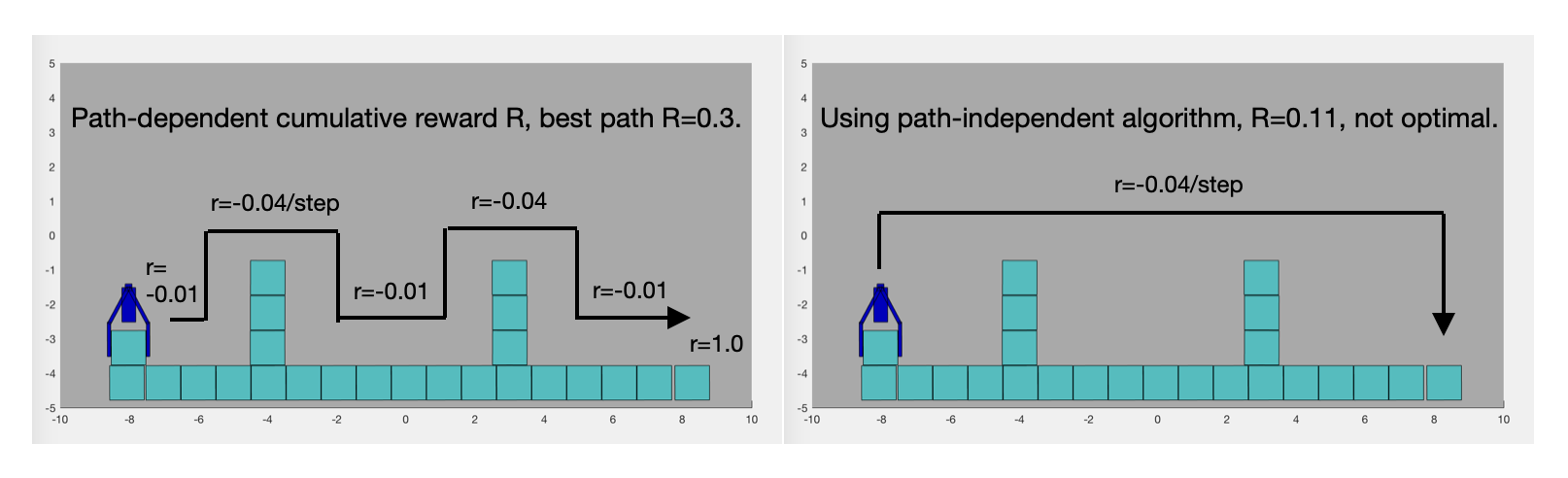}
  \caption{Optimizing algorithm was used on the left to find maximal R terrain following solution. On the right, a feasible solution with no rewards along the way, but R calculated using rewards on left.}
  \label{exp_terrain_follow}
\end{figure}

\begin{figure}
  \centering
  \includegraphics[width=0.6\linewidth]{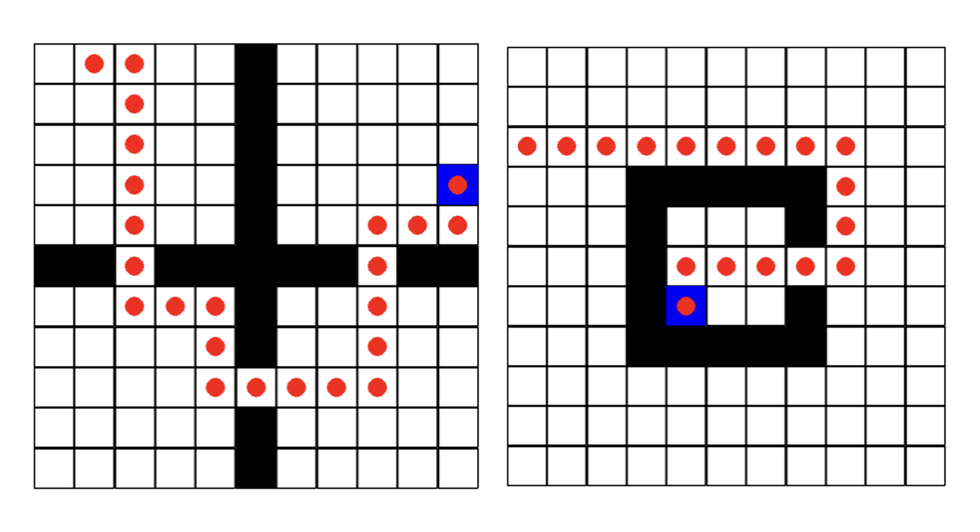}
  \caption{Four rooms using policies learned with original barrier in Figure~\ref{exp_four_rooms} but applied zero-shot to non-original barriers. In this case, the optimizing algorithm was used. It is slow, however, partly because policies were for feasible solutions. Further work is needed to speed these up.}
  \label{exp_four_rooms2}
\end{figure}

\subsubsection{Terrain follow}
\label{terrain_follow}

Figure~\ref{exp_terrain_follow} shows an experiment which tests the optimal path algorithm described in section \ref{TP}. The goal is to move the block from the left-most to right-most block while maximizing the cumulative reward. Negative rewards are given along the way and a positive reward=1.0 is given at the goal. To favor terrain following, the rewards become more negative at higher elevation. The terrain following solution has cumulative reward R=0.3. For comparison, a feasible solution with no rewards along the way, is also shown in Figure~\ref{exp_terrain_follow}. When plugged back into the negative reward problem, its cumulative reward is R=0.11, quite a bit worse than the optimal solution.

\subsubsection{Four rooms}
\label{exp_opt_four_rooms}

As shown in Figure~\ref{exp_four_rooms}, polices learned for the original barrier in the first four panels can be used to find feasible solutions for novel barriers in the last two panels. The algorithm finds these very quickly. On the other hand, if negative rewards are introduced, the optimizing RTP algorithm will find the optimal solutions shown in Figure~\ref{exp_four_rooms2}. Optimizing is done at all 3 levels, but the most valuable optimization turns out to be at levels 2 and 3. Unfortunately, optimizing at these higher levels is slow because many level 1 planners are called and also because the policies were learned from feasible plans only. Re-learning the policies from optimal solutions is planned for future work.

\section{Discussion}
\label{discussion}

I believe this paper demonstrates that a hierarchical tree planner is a good laboratory for studying the benefits of planners for policies and policies for planners. The properties of this planner which turn out to be most helpful are: hierarchy, near-greedy search, boundary states, on-the-fly planner initialization, multiple sub-goal states, and background and object number invariance, although not all of these are important for all problems. I am hoping that at least some of these will be useful in improving other planners too.

One thing that surprised me was that a hierarchical tree planner could solve problems with continuous actions and states and also long horizons, such as the inverted pendulum. On the other hand, the inverted pendulum has low dimensional actions and states, so the next step is to see if hierarchical tree search can solve higher dimensional 3d problems. Another type of problem to investigate is mixed task and motion since an advantage of hierarchical tree search is that it can in principle solve both together.

Policies for all levels of the inverted pendulum and four rooms problems were learned simultaneously, and used natural sub-problems. The lunar lander used a relaxed version of the main problem as a sub-problem. From one point of view, the other experiments also used sub-problems, such as grab and place\_on for block stacking, which are relaxed versions of the main problem, but, in fact, they were engineered and learned separately. Hence, one future goal is to find sub-problems more organically, either as clearly relaxed versions of the main problems, or simultaneously while learning the main problem. 

Another promising area of research is reward engineering for the RTP. This is discussed briefly in Appendix~\ref{appendix_reward_engineering}, but shows great promise in reducing the search tree size and hence search speed down by many orders of magnitude. It is also not as unprincipled as it might sound since the basic idea is to reduce the entropy of the system or the maximum number of states that can be encountered by the robot starting from some initial state. There are many problems where this can be accomplished by placing some sensors on the robot so it doesn’t crash and move around objects that are not relevant to its task. This is ongoing research.

Policies for sub-problems can either be initialized randomly or initialized from some previously learned policies. Actually it is quite possible that even random sets of sub-problems (generalized actions) would be sufficient for higher level planning as long as they sample the state space well enough. At the other extreme, an interesting option is to try an engineered set of developmental sub-problems, for example a set that models Piaget’s theory of sensory motor development \citep{piaget}.

For real robots, a dynamics simulator may be good enough to learn initial policies, but not good enough to robustly plan in a real environment. In this case, PLP policies can be used to initialize robot policies which might then be refined in a real environment using RL, for example. These RL policies could then be used on their own or fed back into the RTP which could propose a near-greedy solution for greater robustness and to help handle novel problems.

\small
\bibliography{bibliography}

\setcounter{equation}{0}

\appendix
\section{Implementation Details}
\label{appendixA}

\numberwithin{equation}{section}
\numberwithin{figure}{section}

\subsection{Boundary states $S_B$}
\label{appendix_boundary_states}

Boundary states $S_B$ are those encountered during search when trying to transition into an obstacle or other forbidden state. These are pretty much the same boundary states added to the search set by RRT (or RDT), see \cite{lavalle} p189-192, chapter 5.5. These boundary states have high sampling value in a hierarchical planner, as will be discussed.

As an example, consider solving the pick and place problem shown in Figure~\ref{exp_pick_place}b using a two level hierarchy. The low level generalized actions are $grab(n)$ and $place()$ where $n$ is the object to grab. Both grab and place were learned with no obstacles, Figure~\ref{exp_grab_place}a,b, so they sometimes bump into the obstacle. When they do so, they return the $S_B$ states near the obstacle, back up to the higher level planner. The higher level planner can then call a few primitive actions to get around the obstacle, and then call $grab$ or $place$ again to finish solving the pick and place problem. This allows the RTP to exploit very simple and low level generalized actions to solve a novel complex problem. The simple generalized actions don’t \textit{see} other objects so they go crashing into them, but that is also why they generalize so well. Thanks to $S_B$ states, however, they can be used to explore the contours of their environment, and to do so more efficiently than a flat primitive action planner.

\subsection{Learning sub-goals}
\label{appendix_sub_goals}

The RTP algorithm solves only one global problem class, formulation \ref{prob_form_1}, at a time for all levels of the RTP hierarchy. The formulation provides the global goals $S_G$. Each sub-planner has its own set of sub-goals $a(g).S_g$ provided by it’s generalized action $a(g)$. The policy for $a(g)$ has typically been previously learned by using a planner to solve the global problem with $S_G=a(g).S_g$. For example, the policy for $grab(n)$, Figure~\ref{exp_grab_place}a, was learned using the global problem with goal states $S_G(grab)$ which are the states where the object ends up in the gripper. Later, when grab is used as a generalized action inside the hierarchy, to solve some other global problem with some other $S_G$, $S_G(grab)$ is no longer available. While it may be possible in some cases to engineer $S_G(grab)$ into $a(g).S_g$, in many cases it is better to learn a reward-type function $r_g(s)$ which is true when $s\in S_G(grab)$. The same search tree data used for policy learning can be used to learn $r_g(s)$, although, unlike the policy, both positive and negative examples are needed. These are both available from the search tree. 

\subsection{Reward engineering}
\label{appendix_reward_engineering}

While the RTP algorithm is domain independent, the problem formulation is not, and there is no reason why rewards and goal states and forbidden states cannot be engineered to improve planning performance when possible. The key engineering principle is to keep the number of states that can be encountered using available actions from each start state $s_I$ low, because the search tree contains at most one node per state. Hence search speed is proportional to the number of encountered states.

To be concrete, consider the place\_in\_box problem with obstacles in Figure~\ref{exp_grab_place}c with primitive actions only. If the gripper is free to knock into any of the objects and move them around and also to drop the object in the gripper, then the search tree for finding a place\_in\_box solution can be enormous. A very crude calculation for the number of possible states is $X^9*X*Y$ where $X$ is the number of discrete horizontal positions that the object or the gripper can occupy and $Y$ is the number of allowed $Y$ positions. There are 9 objects, plus the gripper. With $X=20$ and $Y=10$ for this problem, the maximum tree search size is $20^{10}*10$, which is enormous. A more careful analysis, calculating the number of possible states within a maximum horizon still gives a maximum tree size of over $10^8$ which is not practical.

However, consider what happens if the gripper is not allowed to touch or move objects that are not already in the gripper and is not allowed to drop the object except in the box. For a real robot, these constraints are easy to implement with cheap IR and touch or torque detectors. For the RTP, they are easily engineered as large negative rewards or forbidden states. Then the only set of states to search are the possible positions of the gripper which is approximately $X*Y$ or 200 for this problem. This is easy, so with these constraints even a flat planner can find a way around the obstacles and place the object in the box.

Even though each such problem with constraints can be solved very easily, the set of such problems covers an extremely large set of initial states. Therefore, a policy learned from many such examples can learn a complex and general policy that can avoid obstacles.

Reward engineering can be combined with hierarchy and boundary states to further reduce the search space and make planning possible where it might not be otherwise.

\subsection{Continuous states}
\label{appendix_continuous_states}

Here I discuss how to modify the TP algorithm in Figure~\ref{fig_TP} to accommodate continuous states when there are non-zero rewards along paths to the goal $S_G$. Application to the RTP algorithm is straightforward, but will not be discussed. Also, the modifications here do not guarantee that the TP algorithm will find optimal solutions for continuous states, but they do produce monotonic improvements in most cases although this too is not guaranteed. Future work is planned to fully explore optimality in the continuous case.

As discussed in Section~\ref{continuous}, the equals function $e(s',s'')$ is used in lines 7 and 15 to decide if state $s’$ is already in the tree $T$. Now, suppose $s''$ is the state in the tree with largest current $R(s'')$ of those states for which $e(s’,s'')=true$. Then if, in addition, $R(s)+r>R(s’')$ in line 15, then one should replace the existing path to $s''$ with the one to $s’$. However, in the continuous case one first needs to make sure $s''$ is not already in the path to $s’$ which would introduce a loop. (This ancestor check is also needed for problems with positive rewards, see Appendix~\ref{appendix_positive}) If $s''$ is not an ancestor, and if $R(s)+r>R(s'')$, then the old $s''$ and any sub-tree that starts with $s''$ are simply removed from the tree. The sub-tree attached to $s''$ cannot be used in the continuous case because $s''$ is not precisely equal to $s’$ so the dynamics from $s’$ might be inconsistent with that sub-tree. 

Hence, the algorithm for the continuous case is much simpler, and it will, in each step, replace a lower $R(s'')$ path with a higher $R(s’)$ path. However, in a next step, $s’$ might be replaced by some other $s'''$ which is within sample distance $e(s’,s''')$ of $s’$, and this does not imply that $e(s''',s'')$ is also true. This means that the algorithm can drift and cannot guarantee that the paths found are the optimal ones that pass through all previously sampled regions. Hence the need for more work on the continuous algorithm.

\subsection{Background filtering}
\label{appendix_background_filtering}

Simple low level tasks can be used as building blocks for more difficult tasks. For example, low level $grab(n)$ can be used for pick and place and for block stacking. $grab(n)$ was learned from scenes, Figures~\ref{exp_grab_place}a with just one object, but was then used as a generalized action in scenes with many objects. To make this possible, the policy for grab uses an unlearned pre-filter, $f(s,n)$ (not to be confused with the transfer function), as a front-end to its policy $p(a|s)=p_f(a|f(s,n))$ where $p_f$ is the function that is actually learned. In $f(s,n)$, $n$ is the object label.

As an example, lets suppose that in a scene like Figure~\ref{exp_pick_place}a with $N$ objects, we use the input state represent 

\begin{equation} 
\label{appendix_srep}
s=(x_g,y_g,x_1,y_1,…,x_N,y_N)
\end{equation}

where $(x_g,y_g)$ is the gripper position and $(x_i,y_i)$ is the position of object $i$, and the lossy filter

\begin{equation} 
\label{appendix_filter}
f(s,n)=(x_g,y_g,x_n,y_n) \quad for \; object \; i=n
\end{equation}

When used in scenes with obstacles, for example, a policy with the filter (\ref{appendix_filter}) will not \textit{see} the obstacles, but this is no problem since the planner can use a near-greedy policy or a hierarchy, and use boundary states, as discussed above.

\subsection{Object number invariance}
\label{appendix_number_invariance}

The filter in (\ref{appendix_filter}) has another benefit: the policy becomes invariant with respect to the number of objects in the scene. This means that the policy can be learned from and used in scenes with any number of objects.

On the other hand, policies for more complex problems may not be able to use a filter as simple as the one in (\ref{appendix_filter}). For example, as was discussed in \ref{speed_block_stacking}, a level 2 policy to stack 3 blocks is much easier to learn than one that stacks 5, see Figure~\ref{exp_block_stacking}. This level 2 policy can generalize from 3 to 5 blocks because it is approximately object number invariant. It achieves this by only focussing on local features in the scene.

In greater detail, the 3 block policy $p(a|s)$ has 6 generalized actions $a=\{grab(n),place\_on(m)\}$ where $n,m=0,1,2$ label the 3 blocks, and $grab(n)$ grabs block $n$, while $place\_on(m)$ places the object in the gripper on block $m$. To decide which object $n$ or $m$ to act on next $p(a|s)$ needs only local scene information. For example, it needs to know which object is the top-most block on the current stack. For this local information, a 2d image representation for state $s$ works best together with a neural network such as a \textit{cnn} that focusses on local features. For example, if objects live on a discrete grid $(x,y)$ one can use

\begin{equation} 
\label{appendix_srep2d}
s=I(x,y) \quad with \; I(x_i,y_i)=1 \; for \; object \; at \; (x_i,y_i), \quad I(x,y)=0 \; otherwise
\end{equation}

The choice of which object to $grab$ or $place\_on$ then becomes equivalent to choosing where in the scene, $(x,y)$, the robot should act next. This is facilitated by using a 2d representation for the policy as well: $p(a|s) \rightarrow p(x,y|s)$. Actually, $grab$ and $place\_on$ each need their own such 2d policy, as discussed below in Appendix~\ref{appendix_policy_splitting}. The output of the 2d probability function is a 2d probability map which shows \textit{where} in the scene the robot should act.

The combination of a 2d input representation, a local-feature neural network, and a 2d probability function, produces approximate background and object number invariance because the policy ignores what's happening outside of a local region. It is this approach that allows zero-shot transfer from 3 to 5 blocks.

I should mention, that going from 3 to 5 blocks also requires using the on-the-fly initialization $a.init(s)$, discussed in Section~\ref{GA}, to adjust the number of generalized actions $a.A$. No modification of the 3 block policy is required at all.

\subsection{2d 1-hot actions}
\label{appendix_2d_1_hot_actions}

To learn the 2d policy $p(x,y|s)$ discussed in \ref{appendix_number_invariance}, it turns out that a 2d image-like output which tries to predict a 2d 1-hot target works best. This 2d 1-hot target replaces the 1d 1-hot target $O(a,s)$ used in the cross entropy in Section~\ref{learning_algorithm}. As an example, if the action $a=grab(n)$, then the 2d 1-hot action would have a 1 at location $(x,y)$ of object $n$ and 0s everywhere else. (assuming objects live on a grid.) The neural network that learns $p(x,y|s)$ takes a 2d input $s$ using a representation such as (\ref{appendix_srep2d}). The network uses a 2d soft-max output for $p(x,y|s)$ so that the resulting stochastic policy $p(x,y|s)$ is a probability map showing which location has the object that should be acted on next. This 2d policy output uses the 2d 1-hot action representation as the target $O(a,s)$ in equation (\ref{cross_entropy}). 

In block stacking, for example, if in addition to a generalized action for $grab(n)$, there is one for $place\_on(m)$, we will also need a 2d stochastic policy for $place\_on(m)$, and also another 1d policy to choses between $grab(n)$ and $place_on(m)$. These policy pieces then need to be put together in a principled way into a single policy $p(a|s)$ where the index $a$ runs over $\{grab(n), place\_on(m)\}$ for all possible $n$ and $m$. Appendix \ref{appendix_policy_splitting} describes how to split $a$ into tasks $i$ such as $i=\{grab, place\_on\}$ and sub-tasks $j=\{n,m\}$ and then how to combine them into a single $p(a|s)$.

\subsection{Policy splitting}
\label{appendix_policy_splitting}

Each generalized action, \ref{GA}, $a(g)$ itself has a set of generalized and primitive actions, $a(g).A$ it can call. Let the index $i$ be an index into that set: $a(i) \in a(g).A$. Let’s further assume that the tasks corresponding to $i$ are best grouped into tasks and sub-tasks. To be concrete, let the high level $a(g)$ be a generalized action that solves the pick and place problem, Figure~\ref{exp_pick_place}. It can call generalized actions $a(g).A=\{grab(n),place()\}$ which are lower level GA which grab object $n$ or place the object in the gripper into the open box. If there are 5 objects in the scene, then $i=0-5$ since $|a(g).A|=6$. One could learn a policy $a(g).p(i|s)$ for $a(g)$ using, for example, a neural network with six soft-max outputs which targets a size 6 one-hot vector for $i$. 

However, this neural network would have to work pretty hard to solve two problems which are much easier to solve separately: the first is to decide whether to call $grab$ or $place$ and the second, given that one has decided to call $grab$ is to determine which object $n$ to grab. In fact, deciding between $grab$ and $place$ is almost trivial, the robot will always choose $grab$ if there isn’t an object in the gripper, and always $place$ otherwise. So learning is easier if the index $i$ is split into a task index $j=0,1=\{grab,place\}$ and a sub-task index $k_j$, where $k_0=n=0-4$ ranging over the 5 objects, and $k_1=0$ since place has 1 trivial sub-task (by definition). Putting $p(i|s)$ back together:

\begin{equation} 
\label{appendix_psplit_1}
p(i|s) =  p(j,k_j|s)= p(j|s)*p_j(k_j|j,s)*N(j)
\end{equation}

The reason for the normalization $N(j)$, is that $p(j|s)*p_j(k_j|j,s)$ doesn’t actually obey the chain rule because there can be different numbers of sub-tasks $k_j$ for each task $j$. In the example, $k_j$ has 5 values for $j=0$, $grab$, but only one for $j=1$ $place$. To make a choice for $N(j)$ let’s consider the case where it doesn’t matter which object is grabbed so task $j=0$ has 5 equally likely subtasks, $p_0(k_0|0,s)= 0.2$, while task $j=1$ has only 1 sub-task, $p_1(p_1|1,s)= 1.0$. As a result, $p(j,k_j|s)$ favors tasks with fewer sub-tasks. To compensate, I use the normalization, 

\begin{equation} 
\label{appendix_psplit_2}
N(j)=1/max_{k_j}(p_j(k_j|j,s)) 
\end{equation}

In the above example then, when sub-tasks are equally likely

\begin{equation} 
\label{appendix_psplit_3}
\begin{split}
	p(1,p_1|s)=p(j=1|s)*(0.2)*(1/0.2)=p(j=1|s) \\
	p(2,p_2|s)=p(j=2|s)*(1.0)*(1/1.0)=p(j=2|s) \\
\end{split}
\end{equation}

which intuitively makes sense, since if the sub-tasks are equally likely it shouldn’t matter how many there are when deciding which task $j$ to choose. Another way to see this is that object number invariance is lost if choosing to grab or place depends on the number of objects in the scene.

At the other extreme, when $p_j(k_j|j,s)= 1.0$ for $k_j = 1$, for example, no harm is done because the normalization is 1.0/1.0. This is the case when there is only one object in the scene that it makes sense to grab, but again, this is irrelevant when choosing whether to grab or place.

In practice, in pick and place and block stacking and all such problems, I have found policy splitting makes neural network policy learning much easier and faster and the above normalization works well.

\subsection{Policy invariants}
\label{appendix_policy_invariants}

The filter $f(n,s)$ (\ref{appendix_filter}) used for background filtering has the object label invariance property

\begin{equation} 
\label{appendix_pinvariants_1}
f(n,s)=f(m,s')	\quad s'=s(n \leftrightarrow m)
\end{equation}

where scene s’ is the same as s, except that the object labels n and m are exchanged. This allows a policy learned by acting only on objects with label $n$ to then generalize to objects with any label. It facilitates background and object number invariance too. To obtain other types of policy generalization, the filter can include other invariants. For example, location invariance is obtained by using a filter $f(x,s)$ labeled by location $x$ which has the property that

\begin{equation} 
\label{appendix_pinvariants_2}
f(x,s)=f(x+dx,s') \quad s'=s(x+dx)
\end{equation}

\section{Optimization}
\label{appendixB}

\begin{figure}
  \centering
  \includegraphics[width=0.8\linewidth]{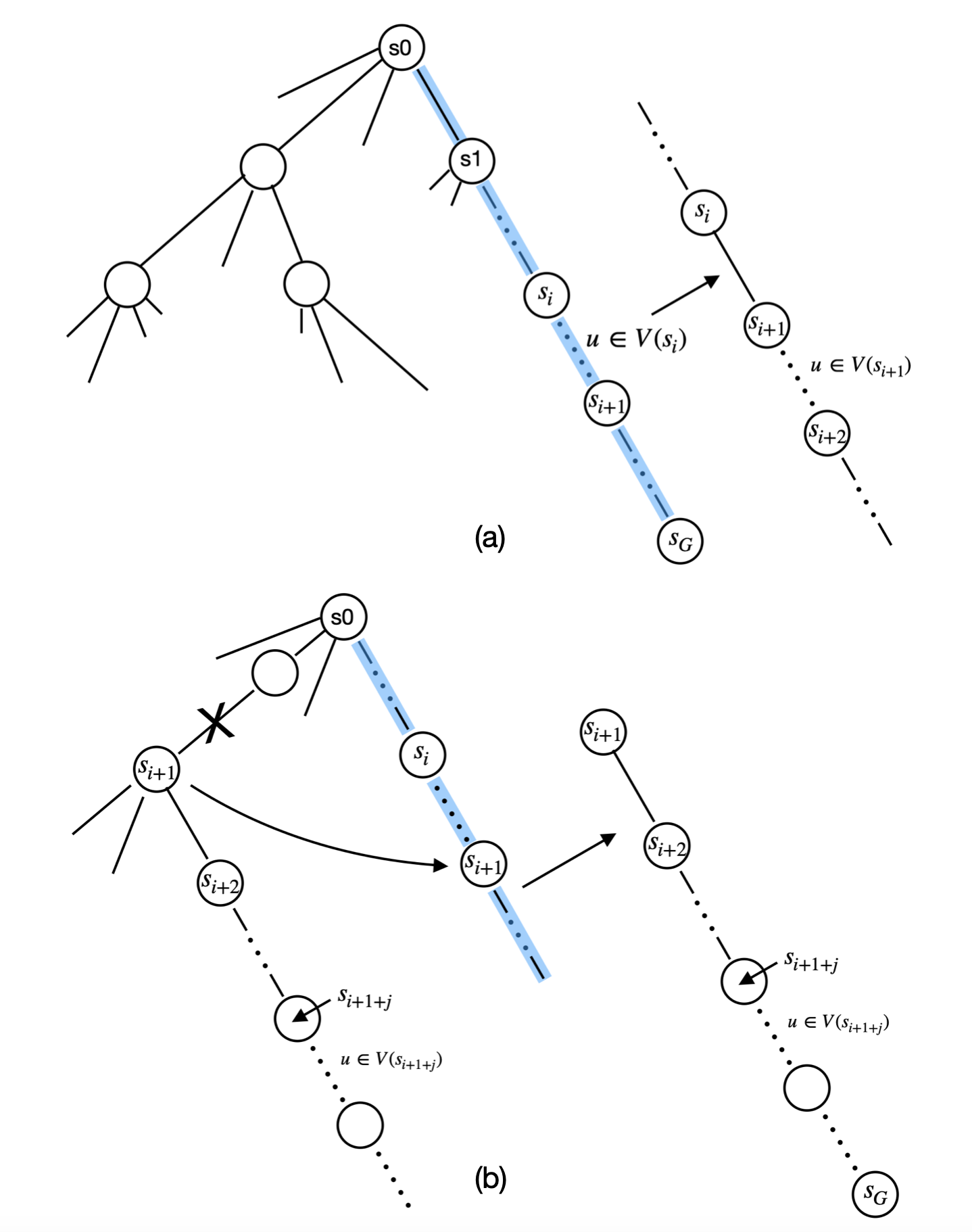}
  \caption{The algorithm constructs the optimal path in blue. (a) Current path to $s_i$ cannot be removed because it is optimal. Next state $s_{i+1}$, will be added eventually because action $u \in V(s_i)$ is available. If $s_{i+1}$ is not in $T$ then it is added and action $u$ is added to $V(s_{i+1})$ which sets up the algorithm to add $s_{i+2}$. (b) If $s_{i+1}$ is already in $T$ then it will be moved to blue path along with sub-tree and all non-edge actions will be refreshed and added to $V(s)$ for all $s$ in the subtree. Therefore, the next state to add is $s_{i+2+j}$ has $u \in V(s_{i+1+j})$ so when this $u$ is tried  $s_{i+2+j}$ will be added.}
  \label{appendix_B}
\end{figure}

\subsection{Informal Proof}
\label{appendix_proof}

The following proof of optimality for the TP algorithm, Figure~\ref{fig_TP}, seems pretty sound but it is not mathematically rigorous. I expect that the algorithm is known and a better proof exists but I have not yet found the references. On the other hand, the informal proof here may be a useful exercise as a way to explore the algorithm. For background see \citep{dijkstra,lavalle}.

To start, I define an optimal path as one that starts at initial state $s_I$ — root node of the search tree — and has the largest cumulative reward $R(s_G)$ to the goal state $s_G$. An optimal sub-path from $s_I$ to $s_i$ has the largest cumulative reward $R(s_i)$ of all paths from $s_I$ to $s_i$.

For the proof, I assume that the optimal path contains no loops: states cannot appear twice on the path. I also assume that the transition function has the Markov property. The Markov property implies that any sub-path from state $s_A$ to state $s_B$ can be substituted for any other sub-path with the same endpoints. Therefore, the cumulative reward of a path is simply the sum of the cumulative rewards of its non-overlapping sub-paths. 

The no-loops assumption together with the Markov property imply that \textit{all sub-paths of the optimal path are also optimal}. To see this, first note that if some sub-path was not optimal, one could simply cut it out and replace it with a more optimal sub-path. The total cumulative reward of the original full path would then be higher, which would mean the original path wasn’t optimal in the first place. 

The exception would be a finite optimal path with loops. Such an optimal path is possible if the path length is constrained, which is a common constraint. (Without the constraint, sub-paths with loops could be added an infinite number of times.) A finite optimal path with loops will contain sub-optimal sub-paths, hence the need to assume, as above, that the optimal path does not contain loops.

The proof constructs the optimal path shown in blue in Figure~\ref{appendix_B}a. If there is more than one optimal path, at least one of them will be constructed. Recall, Figure~\ref{fig_TP}, that $V(s)$ is the set of actions $u$ which are available to expand from state $s$, and $T$ is the search tree with nodes corresponding to states $s$ and edges corresponding to actions $u$.

I assume that the optimal path exists, and has been found up to state $s_i$. Note that this sub-path cannot be removed by the algorithm because it is an optimal sub-path by assumption, so for any $s_j$ on this sub-path, there cannot be any node $s$ in the tree for which $R(s)+r(u,s)>R(s_j)$, line 15 in Figure~\ref{fig_TP}. Line 15 is the only mechanism for removing an existing node $s_j$ from the tree.

Note that the action $u(s_i \rightarrow s_{i+1}) \in V(s_i)$ which means that the action from $s_i$ to $s_{i+1}$ is available. This has to be true, as shown below, because, when $s_i$ was added to the optimal path it was either through line 9 or lines 16-23 in Figure~\ref{fig_TP} and in either case all of it’s non-edge actions were made available ($\in V(s_i)$). Since there is no edge between $s_i$ and $s_{i+1}$ yet, $u(s_i \rightarrow s_{i+1})$ must be available.

The optimal path is constructed as follows:

1. At some point the action $u(s_i \to s_{i+1})$ must be tried, line 4, because it is available, $u(s_i \to s_{i+1}) \in V(s_i)$, and the algorithm tries all available actions.

2. If $s_{i+1} \not\in T$, line 7, it is added to $T$ and $u(s_{i +1} \to s_{i+2})$ is added to $V(s_{i+1})$ because it is included in $U(s_{i+1})$. Set $j=0$. See Figure~\ref{appendix_B}a.

3. If $s_{i+1} \in T$, line 15, then since $s_i \to s_{i+1}$ is on the optimal path the inequality must be true so $s_{i+1}$ is moved to $s_i$ along with its sub-tree if any. (See exception, below.) If this sub-tree includes an edge from $s_{i+1}$ to $s_{i+2}$ then the optimal path is extended to $s_{i+2}$ and so on until an action $u(s_{i+1+j} \to s_{i+j+2})$ is reached for which there is no edge in the sub-tree ($j$ can be $0$). This action is added to $V(s_{i+j+1})$, made available, because all non-edge actions in the sub-tree are revived, line 23. See Figure~\ref{appendix_B}b.

4. If $s_{i+1}=s_G$, stop. Else set $i \leftarrow i+1+j$  and go to 1.

The \textit{exception} in 3. is that the current path to $s_{i+1}$ has $R(s_i)+r(u,s)=R(s_{i+1})$ so $s_{i+1}$ is not moved. But this is simply the case of multiple optimal paths. We continue to assume that $s_{i+1}$ has been moved to the blue path, it’s just a different blue path from $s_0$ to $s_i$, and that upstream portion is now fixed in place so the algorithm continues on as before.

\subsection{Feasible plans}
\label{appendix_feasible}

The exception to step 3, above, is the actually rule for feasible plans where all cumulative rewards $R(s)=0$. In this case, assume that we are on a feasible plan up to $s_i$, then either the next state $s_{i+1}$ on this plan will get added to it, or it will already exist in the tree. If it exists in the tree then the current path to $s_{i+1}$ becomes the new upstream portion of the feasible plan, and we move on to $s_{i+2}$. The point is that it doesn’t matter how we get to any given state, since the transition function is Markov and we are only looking for a feasible plan. Continuing in this way up to the goal state guarantees discovery of a feasible plan assuming one exists.

\subsection{Negative rewards}
\label{appendix_negative}

A sufficient condition for the optimal path to not contain loops is that all non-goal rewards must be negative. This follows because the cumulative reward of a loop will then always be negative. One could therefore simply cutout any loop to increase the cumulative reward, so optimal paths will not contain loops. The negative reward restriction here, is equivalent to the positive costs requirement for the typical Dijkstra algorithm \citep{lavalle} because rewards are maximized while costs are minimized.

Moreover, if rewards are all negative then the algorithm will never produce loops inside the search tree. To see this, note that the only way a loop can be created is on line 15 when the state $s’$ is already in the tree and $s'$ is also on the path leading to $s$: $s'$ is an ancestor of $s$. Now, if all rewards are negative and $s'$ is an ancestor of $s$ then $R(s)<R(s')$. In addition, $R(s)+r(u,s)<R(s)$ because all $r(u,s)$ are negative. Putting these together gives $R(s)+r(u,s)<R(s')$. This means that the inequality in line 15 cannot be true and a loop cannot be created.

An example of a problem with negative rewards is the checkerboard in Figure~\ref{checkerboards}a with random negative rewards. In this case, any order for choosing next actions and states to expand produces the same optimal solution.

\subsection{Positive rewards}
\label{appendix_positive}

When rewards are positive or mixed then the algorithm can introduce loops. This is not a good thing because with loops the algorithm may not converge. More significantly, when $s'$ line 15 is already on the path leading to $s$, the algorithm will cut off the tail of the path which detaches it from the initial state $s_I$. Loops can be avoided by adding an ancestor-detector on line 15, although this slows down the algorithm a little. When the ancestor-detector discovers that $s'$ is already in the path to $s$ no change is made to the tree. With positive or mixed rewards, assuming it has an ancestor-detector, the algorithm still monotonically finds paths with higher cumulative reward. However, optimal solutions are not guaranteed.

An example of a problem with positive rewards is shown in Figure~\ref{checkerboards}b where the rewards are uniform and positive. With an ancestor detector added to line 15, and a uniform search the optimal solution shown was found. However, random search finds good, but not optimal solutions.

\end{document}